\documentclass[11pt]{article}
\usepackage{amsmath}
\usepackage{graphicx}
\usepackage{enumerate}
\usepackage{algorithm}
\usepackage{algpseudocode}
\usepackage{natbib}
\usepackage{url} % not crucial - just used below for the URL 

\usepackage{multirow}
\usepackage[colorlinks=false,allbordercolors={1 1 1}]{hyperref}
\usepackage{booktabs}
\usepackage{amsfonts}
\usepackage{float}
\usepackage{xcolor}

\usepackage{enumitem}
\newlist{steps}{enumerate}{1}
\setlist[steps, 1]{leftmargin=*, label = Step \arabic*:}

\usepackage[top=1in, bottom=0.75in, left=1in, right=1in]{geometry}

\newtheorem{assumption}{\sc Assumption}
 
\newtheorem{theorem}{Theorem}
 
\newtheorem{proposition}[theorem]{Proposition} 
\newtheorem{remark}[theorem]{Remark}
\newtheorem{corollary}[theorem]{Corollary}

\newcommand{\bftab}{\fontseries{b}\selectfont}
\newcommand{\comment}[1]{}

\newcommand{\BlackBox}{\rule{1.5ex}{1.5ex}}  % end of proof
\ifdefined\proof
    \renewenvironment{proof}{\par\noindent{\bf Proof\ }}{\hfill\BlackBox\\[2mm]}
\else
    \newenvironment{proof}{\par\noindent{\bf Proof\ }}{\hfill\BlackBox\\[2mm]}
\fi

\begin{document}

%\bibliographystyle{natbib}

%%%%%%%%%%%%%%%%%%%%%%%%%%%%%%%%%%%%%%%%%%%%%%%%%%%%%%%%%%%%%%%%%%%%%%%%%%%%%%

\def\spacingset#1{\renewcommand{\baselinestretch}%
{#1}\small\normalsize} \spacingset{1}

\title{\bf Shapley Curves: A Smoothing Perspective}
\author{Ratmir Miftachov\thanks{E-mail: ratmir.miftachov@hu-berlin.de. School of Business and Economics and Institute of Mathematics, Humboldt University of Berlin, Germany}\and Georg Keilbar\thanks{E-mail: georg.keilbar@hu-berlin.de. Chair of Statistics, Humboldt University of Berlin, Germany}\and Wolfgang Karl Härdle\thanks{E-mail: haerdle@hu-berlin.de. School of Business and Economics and BRC Blockchain Research Center, Humboldt University of Berlin, Germany;
Sim Kee Boon Institute, Singapore Management University, Singapore; 
Dept. Information Management and Finance, National Yang Ming Chiao Tung University, Hsinchu, Taiwan, ROC; Dept. Mathematics and Physics, Charles University, Prague, Czech Republic; 
Institute for Digital Assets, Romania}}  
\maketitle
\bigskip
\begin{abstract}
This paper fills the limited statistical understanding of Shapley values as a variable importance measure from a nonparametric (or smoothing) perspective. We introduce population-level \textit{Shapley curves} to measure the true variable importance, determined by the conditional expectation function and the distribution of covariates. Having defined the estimand, we derive minimax convergence rates and asymptotic normality under general conditions for the two leading estimation strategies. For finite sample inference, we propose a novel version of the wild bootstrap procedure tailored for capturing lower-order terms in the estimation of Shapley curves.
Numerical studies confirm our theoretical findings, and an empirical application analyzes the determining factors of vehicle prices.
\end{abstract}

\noindent%
\vfill

\spacingset{1.5}

\newpage

%Vorherige Danksagung:
%\thanks{
%    The authors gratefully acknowledge the \textit{Deutsche Forschungsgemeinschaft via IRTG 1792 “High Dimensional Nonstationary Time Series”, Humboldt-Universität zu Berlin; the European Union’s Horizon 2020 research and innovation program “FIN-TECH: A Financial supervision and Technology compliance training programme”under the grant agreement no. 825215 (Topic: ICT-35-2018, Type of action: CSA); the European Cooperation in Science \& Technology COST Action grant CA19130 - Fintech and Artificial Intelligence in Finance - Towards a transparent financial industry; the Yushan Scholar Program of Taiwan; and the Czech Science Foundation’s grant no. 19-28231X / CAS: XDA 23020303.}}\hspace{.2cm}

\newpage
\spacingset{1.8}

\section{Introduction}\label{section:introduction}
%Flow: OK.
Modern data science techniques, such as deep neural networks and ensemble methods like gradient boosting, are renowned for their high predictive accuracy. However, these methods often operate as \textit{black boxes}, offering limited interpretability. To address this challenge, the \textit{Shapley value} has gained popularity as a measure of variable importance in recent years. Originally introduced in game theory, Shapley values provide a unique solution to cooperative games \citep{shapley1953value}. In the realm of machine learning, they enable model explanation by quantifying the contribution of each variable to the prediction. This is achieved by calculating the difference between a prediction for a subset of variables and the same subset additional with the variable of interest. This difference is then weighted and averaged over all possible subset combinations, yielding the respective Shapley values. For example, \citet{lundberg2017unified} assume variable independence and introduce an approximation method known as KernelSHAP, which aims to explain the predictions of the conditional mean. Other researchers have proposed variants of Shapley values based on different predictiveness measures, such as the variance of the conditional mean \citep{owen2017shapley, benard2022shaff}.

%Flow: OK.
A major portion of the existing literature is predominantly focused on practical approximations of Shapley values, given that their computational complexity grows exponentially with the number of variables {\citep{chen2023algorithms}. 
However, the theoretical understanding of Shapley values, especially concerning uncertainty quantification, remains somewhat limited. It is crucial to distinguish between uncertainty due to the computational approximation on the one hand and estimation uncertainty on the other. To address the former, \citet{covert2021improving} study the asymptotic properties of KernelSHAP, a weighted least squares approximation for Shapley values. While these findings confirm the effectiveness of the approximation, their relevance for statistical inference is unclear, as they offer no insight into estimation uncertainty.
Another strand of the literature is concerned with estimation uncertainty, which is the key interest of our paper. \citet{fryer2020shapley} focus on estimation uncertainty for the Shapley values by assuming an underlying linear model, hence the Shapley values are defined as a function of the coefficient of determination, $R^2$, instead of the conditional mean. \citet{johnsen2021inferring} propose bootstrap inference for the SAGE estimator, which is a global measure, but do not prove its asymptotic validity.
Additionally, there are other closely related publications from a statistical standpoint that include asymptotic arguments. \citet{williamson2020efficient} define \textit{Shapley effects} as a global measure of predictiveness and prove their asymptotic normality. \citet{benard2022shaff} establish consistency for Shapley effects, utilizing the variance of the conditional mean with a random forest estimator. While these two papers are similar in spirit, they differ from our analysis, as they focus on global measures for variable importance based on predictive measures. In contrast, our research is centered on providing asymptotics for local measures within the conditional mean framework of Shapley values.

%Flow: OK

The purpose of this paper is to conduct a rigorous asymptotic analysis of the Shapley values, adopting a fully nonparametric approach. In doing so, we establish both consistency results and asymptotic normality. It is crucial to first define the estimand at the population level, which we call the (population level) \emph{Shapley curves}. These are $d$-dimensional functions characterizing the importance of a certain value at any point in the support of the covariates. Shapley curves are uniquely determined by the true conditional expectation function and by the joint distribution of covariates. The perspective in this paper is therefore fundamentally different from most existing work on Shapley values specifically, and from variable importance measures more broadly. Rather than merely `explain' a prediction, our goal is to precisely estimate the true (population level) variable importance.

%Flow: OK
To study the asymptotics of Shapley curves, we analyze two types of plug-in estimators previously discussed in the literature.
First, the \emph{component-based} approach requires the direct estimation of all components in the Shapley decomposition. This is achieved by having separate regression equations for all subsets of variables (see e.g., \citet{vstrumbelj2009explaining} and \citet{williamson2020efficient}). Second, the \emph{integration-based} approach requires only a single estimate of the full regression model. The estimates of the lower-dimensional components in the Shapley decomposition are obtained by integrating out the variables not in the given subset. Our asymptotic analysis reveals that this latter approach is closely related to the literature on the marginal integration estimator of additive models \citep{tjostheim1994nonparametric, linton1995kernel}. However, in the Shapley context, this idea has been explored by \citet{frye2020shapley}, \citet{aas2021explaining}, \citet{covert2020understanding}, and \citet{chen2023algorithms}, among others. By relying on asymptotic arguments proven in the nonparametric literature, we bridge the existing gap of asymptotic developments of this estimator within the field of explainability. In the case of dependent features, the integration-based approach relies on an estimate of the conditional densities of the variables. Various estimation strategies, such as the assumption of Gaussian distributions or the application of copula methods, are discussed by \citet{aas2021explaining1}. Additionally, \citet{lin2023robustness} present theoretical findings on the robustness of the integration-based approach concerning variable omission.

%Flow: OK
As we consider a fully nonparametric model setup with general assumptions, we rely on local linear estimation \citep{fan1993local}. We demonstrate that both the component-based and integration-based approaches to estimation achieve the minimax rate of convergence. Furthermore, we establish the asymptotic normality of these estimates. Notably, we prove that the asymptotic distributions of the two estimators differ only in their bias, not in their asymptotic variance. Specifically, the integration-based approach has a larger bias. This finding is not unique to local linear estimation; reliance on a $d$-dimensional pilot estimator will typically lead to oversmoothing of the lower-dimensional components. This oversmoothing effect has implications for other modern smoothing techniques that rely on hyperparameters to balance the bias-variance trade-off, including regression tree ensembles and neural networks. Similarly, using the integration-based approach will result in an inflated bias.

%Flow: OK
While inference based on the asymptotic normal distribution is possible, it often leads to unsatisfactory performance in finite sample scenarios. To achieve better coverage in finite samples, we introduce a consistent wild bootstrap procedure \citep{mammen1992bootstrap,hardle1993comparing} specifically designed for the construction of Shapley curves. To the best of our knowledge, our wild bootstrap procedure, also referred to as the multiplier bootstrap, is the first in the context of estimation uncertainty for local Shapley measures that is proven to be consistent. By generating bootstrap versions of the lower-order terms, we effectively mimic the variance of the estimator counterpart. An extensive simulations study confirms our theoretical results, highlighting the strong coverage performance of our bootstrap procedure.

%Flow: OK
Our contributions to the field are two-fold. The first contribution is of a conceptual nature. By considering population-level Shapley curves we argue for a perspective that is different than the majority of the existing literature. Instead of merely viewing Shapley values as an `explanation' of a given prediction method \citep{lundberg2017unified, covert2020understanding}, our goal is to provide accurate estimates of the true variable importance as determined by the distribution of the data. This is a prerequisite for our second, more technical contribution, which is the rigorous statistical analysis of the two predominant estimation techniques for Shapley curves. By proving that both estimators achieve the minimax rate of convergence in the class of smooth functions and establishing their asymptotic normality, we enhance the theoretical understanding of Shapley values.

%Flow: OK
This paper contributes to the rapidly growing body of literature on variable importance measures. A general overview of state-of-the-art algorithms to estimate the Shapley value variable attributions is given in \citet{covert2021explaining} and, more recently, in \citet{chen2023algorithms}. Further research has focused on improving the computational efficiency of Shapley value estimation, employing strategies such as multilinear extension techniques \citep{okhrati2021multilinear} and FastSHAP \citep{jethani2022fastshap}. Unlike local methods that explain predictions at the level of individual observations, there is also considerable interest in global explanations that apply to the entire sample, such as SAGE \citep{covert2020understanding} and Shapley Effects \citep{owen2014sobol, song2016shapley, owen2017shapley}. Other methods provide algorithm-specific approximations instead of model agnostic, such as decision tree ensembles \citep{lundberg2020local, muschalik2024beyond}.

%Flow: OK
The structure of this paper is as follows. First, we introduce the nonparametric setting and notation of our work in Section \ref{section:population}. The Shapley curves are defined on population level and examples are given for building the intuition of the reader. In Section \ref{section:estimation} we propose two estimation approaches for Shapley curves. Our main theorems regarding asymptotics are given and we elaborate on their heuristics. Additionally, we detail the novel implementation of our wild bootstrap methodology. In Section \ref{section:numerical}, through simulations, we demonstrate that prediction accuracy improves with increasing sample size, and our wild bootstrap procedure achieves good coverage. Section \ref{section:application} applies our methodology to estimate Shapley curves for vehicle prices based on vehicle characteristics. The estimated confidence intervals for the Shapley curves enable practitioners to conduct statistical inference. The paper concludes with Section \ref{section:conclusion}, summarizing our findings and contributions.

\section{A Smoothing Perspective on Shapley Values}
\label{section:population}
%RM:checked
\subsection{Model Setup and Notation}
Consider a vector of covariates $X=(X_1,\ldots,X_d)^{\top}\in\mathcal{X}\subseteq\mathbb{R}^{d}$ and a scalar response variable $Y\in\mathcal{Y}\subseteq\mathbb{R}$. Let $F$ denote the cumulative distribution function (cdf) of $X$ with continuous density (pdf), $f$. Let $\{(X_i,Y_i)\}_{i=1}^{n}$ be a sample drawn from a joint distribution function $F_{X,Y}$. Consider the following nonparametric regression setting,
\begin{align}\label{eq:dgp}
    Y_i=m(X_i)+\varepsilon_i,\quad i=1,\ldots,n,
\end{align}
with ${\sf E}(\varepsilon_i|X_i)=0$ and $m\in\mathcal{M}$, where $\mathcal{M}$ is a rich class of functions. Consequently, $m(x)={\sf E}(Y|X=x)$ is the conditional expectation.
Let $N\stackrel{\text{def}}{=}\{1,\ldots,d\}$, and let $\mathcal{S}$ denote the power set of $N$. For a set $s\in\mathcal{S}$, $X_s$ denotes a vector consisting of elements of $X$ with indices in $s$. Correspondingly, $X_{-s}$ denotes a vector consisting of elements of $X$ not in $s$. We write $m_s\in\mathcal{M}_s$ to denote functions which ignore arguments with index not in $s$, $\mathcal{M}_s=\{m\in\mathcal{M}:m(u)=m(v)\text{ for all }u,v\in\mathcal{X}\text{ satisfying }u_s=v_s\}$. Similarly, we write $m_{-s}\in\mathcal{M}_{-s}$ for functions which ignore arguments in $s$. Finally, we write $f_{X_{-s}|X_{s}}(x_{-s}|x_s)$ for the conditional density functions of $X_{-s}$ given $X_s=x_s$.
The indicator function $\mathbf{I}\{j \in s\}$ is equal to $1$ if $j\in s$ and $0$ otherwise. The sign function $\text{sgn}\{j \in s\}$ takes the value $1$ if $j\in s$ and $-1$ otherwise.

\subsection{Population Shapley Curves}
We now define the population-level Shapley curves, which are functions, $\phi_j(\cdot):\mathbb{R}^{d}\to\mathbb{R}$ measuring the local variable importance of a variable $j$ at a given point $x\in\mathbb{R}^d$. They are uniquely determined by $m(x)$ defined in (\ref{eq:dgp}) and the joint distribution function of the covariates, $F$. Our focus differs from the prevalent perspective in the literature, as we aim not merely to interpret a given prediction but rather to estimate and make inference on the true Shapley curves. As a consequence, looking at variable importance on the population level is agnostic of the specific form a corresponding estimator might take. 

Originally proposed in the game theory framework, Shapley values measure the difference of the resulting payoff for a coalition of players and the same coalition including an additional player \citep{shapley1953value}. By keeping a player fixed, this difference is calculated across all possible coalitions of players and averaged with a combinatorial weight.
From a statistical point of view, each player is represented by a variable and the payoff is therefore a measure of contribution for the corresponding subset of variables. In this work, this measure is set to be the conditional mean. 
Finally, let us define $\phi_j(x):\mathbb{R}^d\to\mathbb{R}$, as follows,
\begin{align}\label{eq:shapley}
    \phi_j(x)=\sum_{s\subseteq N\setminus{j}}\frac{1}{d}\binom{d-1}{|s|}^{-1}\left\{m_{s\bigcup{j}}(x_{s\cup{j}})-m_{s}(x_s)\right\},
\end{align}
for $j\in N$, where the components are defined as
\begin{align}\label{eq:component}
\begin{split}
    m_s(x_s)&=\int m(x)f_{X_{-s}|X_s}\left(x_{-s}|x_s\right)dx_{-s}\\
    &={\sf E}\left(Y|X_s=x_s\right),
\end{split}
\end{align}
for $s\in\mathcal{S}$. Note that $m_N(x)=m(x)$ represents the full nonparametric model and $m_{\emptyset}={\sf E}(Y)$ is the unconditional mean of the response variable. A point $\phi_j(x)$ in the Shapley curve measures the difference in the conditional mean of $Y$ from including the variable $X_j$, averaged through the combinatorial weight over all possible subsets. 

%On Assumptions
The Shapley decomposition satisfies several convenient properties, namely \textit{additivity, symmetry, null feature} and \textit{linearity}. These properties have been initially proved for cooperative games and subsequently transferred to feature attribution methods.
Most importantly, our definition of Shapley curves (\ref{eq:shapley}) satisfies the crucial \textit{additivity} property, $m(x)-{\sf E}(Y)=\sum_{j=1}^{d}\phi_j(x)$.
Namely, the conditional expectation of the full nonparametric model (\ref{eq:dgp}) subtracted by the unconditional mean of the dependent variable, $m(x)-{\sf E}(Y)$, can be recovered exactly as a sum of Shapley curves of variables $X_j$ evaluated at the point $x$. \citet{sundararajan2020many} discuss the remaining properties for the conditional expectation case.

%RM:OK
%Local vs global
In contrast to the Shapley-based variable importance measures of \citet{williamson2020efficient} and \citet{benard2022shaff}, Shapley curves provide a local instead of a global assessment of the importance of a variable. Consequently, the Shapley curve will take different values for different points on the $d$-dimensional support $\mathcal{X}$ of the covariates. It might be the case that a variable is redundant in certain areas of $\mathcal{X}$, but indispensable in other areas. Local measures are thus able to paint a more nuanced picture. 

%RM:OK
We want to highlight the important role that the dependency in $X$ plays in (\ref{eq:component}), and consequently in the definition of the Shapley curves. For a set $s\in\mathcal{S}$, $m_s(x_s)$ represents the conditional expected value of the dependent variable, with variables not in $s$ integrated out with respect to the conditional density $f_{X_{-s}|X_s}(x_{-s}|x_s)$. 
In the special case of independent covariates, the expression for the conditional density simplifies to a product of (unconditional) marginal densities. In general, however, the component (\ref{eq:component}), and consequently $\phi_j(x)$, will depend crucially on the dependency structure of $X$. In particular, the difference in the conditional expected value for a given set $s$, $m_{s\bigcup{j}}(x_{s\cup{j}})-m_{s}(x_s)$, can be a result of the direct effect of variable $X_j$ on $Y$ via the functional relationship described by $m(\cdot)$, or it might be due to the dependence of $X_j$ with another variable $X_k$, which in turn has a direct effect on $Y$. This can be seen as a bias caused by the endogeneity of $X_j$. It is therefore crucial to understand, that Shapley curves, even if they are defined on the population, are a predictive measure of variable importance and not a causal measure. We will demonstrate the role of dependence in a few examples in the next subsection.

\subsection{Examples}

Let us have a look at Shapley curves in a few interesting scenarios. We consider different settings for the regression function, $m(x)$, as well as for the dependence structure of $X$. In particular, we are interested in the difference between the case of independent and dependent regressors.
\subsubsection*{Example 1: Additive Interaction Model}
We first consider the following additive interaction model, with mean-dependent covariates $X_1$ and $X_2$,
\begin{align*}
    m(x_1,x_2)=g_1(x_1)+g_2(x_2)+g_{12}(x_1,x_2).
\end{align*}
This model was discussed in \cite{sperlich2002nonparametric}, \cite{chastaing2012generalized}, and more recently in \cite{hiabu2020random} in the context of a random forest variant. It is used thoroughly in our simulation studies. The Shapley curve on population level for variable $X_1$ is given by
\begin{align*}
    % \phi_1(x)&=\frac{1}{2}\left[g_1(x_1)+g_2(x_2)+g_{12}(x_1,x_2)\right]-\frac{1}{2}\left[{\sf E}\left\{g_1\left(X_1\right)|X_2=x_2\right\}+g_2(x_2)+{\sf E}\left\{g_{12}\left(X_1,x_2\right)|X_2=x_2\right\}\right] \\
    % &+\frac{1}{2}\left[g_1(x_1)+{\sf E}\left\{g_2\left(X_2\right)|X_1=x_1\right\}+{\sf E}\left\{g_{12}\left(x_1,X_2\right)|X_{1}=x_1\right\}\right]-\frac{1}{2}{\sf E}\left(Y\right)\\
    % &=\frac{1}{2}\left[g_1(x_1)-{\sf E}\left\{g_1(X_1)|X_2=x_2\right\}+g_{12}(x_1,x_2)-{\sf E}\left\{g_{12}(X_1,x_2)|X_2=x_2\right\}\right]\\
    % &+\frac{1}{2}\left[g_1(x_1)+{\sf E}\left\{g_2\left(X_2\right)|X_1=x_1\right\}+{\sf E}\left\{g_{12}\left(x_1,X_2\right)|X_{1}=x_1\right\}\right]-\frac{1}{2}{\sf E}\left(Y\right)\\
    \phi_1(x)&=g_1(x_1)+\frac{1}{2}\left[{\sf E}\left\{g_2\left(X_2\right)|X_1=x_1\right\}-{\sf E}\left\{g_1\left(X_1\right)|X_2=x_2\right\}\right]\\
    &+\frac{1}{2}\left[g_{12}(x_1,x_2)-{\sf E}\left\{g_{12}\left(X_1,x_2\right)|X_{2}=x_2\right\}+{\sf E}\left\{g_{12}\left(x_1,X_2\right)|X_{1}=x_1\right\}\right]-\frac{1}{2}{\sf E}(Y).
\end{align*}
By symmetry, $\phi_2(x)$ can be defined in a similar way. It is important to understand how the Shapley curve is affected (i) by the interaction effect, and (ii) by the dependence across the covariates.
In the case of non-zero interaction but under mean independence of covariates, and under the following identification assumptions, ${\sf E}\{g_1(x_1)\}={\sf E}\{g_2(x_2)\}={\sf E}\{g_{12}(x)\}=0$, the expression simplifies to
    \begin{align*}
        \phi_1(x)=g_1(x_1)+\frac{1}{2}g_{12}(x).
    \end{align*}
The simplified expression consists of the main effect and variable $X_1$'s share of the interaction effect. On the other hand, assuming zero interaction, i.e. $g_{12}(x)=0$, but allowing for mean-dependent covariates, we can isolate the effect of dependence,
    \begin{align*}
        \phi_1(x)=g_1(x_1)+\frac{1}{2}\left[{\sf E}\left\{g_2\left(X_2\right)|X_1=x_1\right\}-{\sf E}\left\{g_1\left(X_1\right)|X_2=x_2\right\}\right].
    \end{align*}
Finally, in the absence of any interaction and dependence, the Shapley curve simplifies to the partial dependence function of an additive model, i.e. $\phi_1(x)=g_1(x_1)$.

\subsubsection*{Example 2: Threshold Regression Model}

In our second example, we consider the threshold regression model \citep{dagenais1969threshold}, a well-known econometric model in which the effects of the variables enter non-linearly. A notable economic application is the multiple equilibria growth model \citep{durlauf1995multiple,hansen2000sample}. Empirical analyses suggest the existence of a regime change in GDP growth depending on whether the initial endowment of a country exceeds a certain threshold. More precisely, take the conditional expectation of $Y$ given $x$ as
\begin{align*}
    m(x)=\left\{\psi+\theta\mathbf{I}(x_{2}\leq C)\right\}x_{1},
\end{align*}
where $C,\psi,\theta$ are scalar parameters. Under the assumption of mean-independence among $X_1$ and $X_2$, the population-level Shapley curves for both variables are
\begin{align*}
    \phi_1(x)&=\left\{\psi+\frac{1}{2}\theta\mathbf{I}(x_{2}\leq C)+\frac{1}{2}\theta F_{X_2}(C)\right\}\left\{x_{1}-{\sf E}(X_{1})\right\},\\
    \phi_2(x)&=\frac{1}{2}\theta\left\{\textbf{I}(x_{2}\leq C)-F_{X_2}(C)\right\}\left\{x_1+{\sf E}(X_1)\right\},
\end{align*}
where $F_{X_2}$ is the marginal cdf of $X_2$. The Shapley curve for variable $X_1$ will take a larger value in absolute values whenever the coordinate $x_1$ is far away from the unconditional expectation, ${\sf E}(X_1)$. Similarly, the Shapley curve for the threshold variable $X_2$ will take a large value if the difference $\left\{\textbf{I}(x_{2}\leq C)-F_{X_2}(C)\right\}$ is large, i.e., in situations in which the effect of the variable is not well captured by its unconditional expectation.

\section{Estimators and Asymptotics} 
\label{section:estimation}

In this section, we discuss the two common approaches for estimating Shapley curves. By considering \textit{curves} instead of a \textit{value} of importance, we gain insights on the whole support of the covariates. The goal is to establish consistent and asymptotically normal estimation of these curves in a general nonparametric setting. This is crucial because otherwise, we are not able to determine whether the estimate on the sample level is meaningful in any way. Indeed, \citet{scornet2023trees} and \citet{benard2022mean} demonstrate that a variety of measures for variable importance in random forests only have a meaningful population-level target in the restrictive case of independence of regressors and no interactions. Both approaches discussed in this chapter are plug-in estimators using the Shapley formula in (\ref{eq:shapley}). The first approach is based on separate estimators of all component functions, for all subsets $s$. We call this the \emph{component-based} approach. Alternatively, one can obtain estimators of the component $m_s(x_s)$ using a pilot estimator for the full model, i.e., an estimator of $m(x)$, and integrate out the variables not contained in the set $s$ with respect to the estimated conditional densities using (\ref{eq:component}). This method we denote as the \emph{integration-based} approach. 

\subsection{Component-Based Approach}
\label{comp_chapter}

%basic idea of the component-based approach 
    
The component-based approach for estimating Shapley curves involves the separate estimation of each component. For each subset $s\in\mathcal{S}$, we use a nonparametric smoothing technique to regress $Y_i$ on $X_{s,i}$, i.e., those regressors contained in $s$. This yields in total $2^d$ regression equations,
\begin{align}
    Y_i=m_{s}(X_{s,i})+\varepsilon_{s,i},\quad i=1,\ldots,n;\quad s\in\mathcal{S},
\end{align}
with ${\sf E}(\varepsilon_{s,i}|X_{s,i})=0$ and $m_{s}$ is defined as in equation (\ref{eq:component}). Given an estimator for the components, we obtain a plug-in estimator for the Shapley curve of variable $j$ using (\ref{eq:shapley}). As a result, the estimated Shapley curve follows as
\begin{align}\label{eq:shapley_est}
    \widehat{\phi}_j(x)=\sum_{s\subseteq N\setminus{j}}\frac{1}{d}\binom{d-1}{|s|}^{-1}\left\{\widehat{m}_{s\bigcup{j}}(x_{s\cup{j}})-\widehat{m}_{s}(x_s)\right\}.
\end{align}
Since $m_s$ is not specified parametrically, we employ local linear estimators \citep{fan1993local} for the component functions. Let $Y=(Y_1,\ldots,Y_n)^\top$, $Z_s=(z_{s1},\ldots,z_{sn})^\top$, $z_{si}=(1,X_{s_1,i},\ldots,X_{s_d,i})$ and $K_s=\operatorname{diag}[\{h_s^{-d_s}\prod_{j=1}^{d_s}k(h_{s}^{-1}(X_{s_j,i}-x_j))\}_{i=1}^n]$, where $k$ is a one-dimensional kernel function and $h_s$ is the bandwidth. Then we have,
\begin{align}
    \widehat{\beta}_s(x_s)=\begin{pmatrix}\widehat{\beta}_{s,0}(x_s)\\\widehat{\beta}_{s,1}(x_s)\end{pmatrix}=\left(Z_s^\top K_sZ_s\right)^{-1}Z_s^\top K_sY,
\end{align}
and the local linear estimator is $\widehat{m}_s(x_s)=\widehat{\beta}_{s,0}(x_s)$.

We are interested in the asymptotic behavior in terms of the global rate of convergence and Gaussianity. This has at least two important implications. First, it demonstrates that the estimation of Shapley curves on the sample level targets the correct quantities. Second, it allows us to construct confidence intervals and conduct hypothesis tests on the estimated curves. For this purpose, we impose the following regularity assumptions.

%07.02.24 assumptions:
%\begin{assumption}\label{assum:density}\
%\begin{enumerate}[label=(\roman*)]
%    \item The support of $X$ is $\mathcal{X}$.
%    \item The density $f$ of $X$ is bounded, bounded away from zero and twice continuously differentiable on $\mathcal{X}$.
%    \item ${\sf Var}(\varepsilon|x)=\sigma^2(x)<\infty$ for all $x\in\mathcal{X}$.
%    \item ${\sf E}(|Y|^{2 + \delta} \lvert X=x)< \infty$ for some $\delta > 0$.
%\end{enumerate}
%\end{assumption}

%Um platz zu sparen:
\begin{assumption}\label{assum:density}\
(i) The support of $X$ is $\mathcal{X}$. (ii) The density $f$ of $X$ is bounded, bounded away from zero, and twice continuously differentiable on $\mathcal{X}$. (iii) ${\sf Var}(\varepsilon|x)=\sigma^2(x)<\infty$ for all $x\in\mathcal{X}$. (iv) ${\sf E}(|Y|^{2 + \delta} \lvert X=x)< \infty$ for some $\delta > 0$.
\end{assumption}

\begin{assumption}\label{assum:function}
Assume $m(x)$ belongs to $\mathcal{M}_d$, the space of $d$-dimensional twice continuously differentiable functions.
\end{assumption}

\begin{assumption}\label{assum:kernel}
Assume $k(\cdot)$ is a univariate twice continuously differentiable probability density function symmetric about zero and $\int s^2k(s)ds=\mu_2(k)<\infty$ and $\int k^{2+\delta}(s)ds<\infty$ for some $\delta>0$.
\end{assumption}

The following theorem shows the consistency of the component-based estimator in the mean integrated squared error (MISE) sense.
\begin{proposition}
\label{proposition:conv_comp}
Let $\widehat{\phi}_j(x)$ be the component-based estimator with components estimated via the local linear method with bandwidths $h_s\sim n^{-\frac{1}{4+|s|}}$. Then we have under Assumptions \ref{assum:density}, \ref{assum:function} and \ref{assum:kernel}, as $n$ goes to infinity,
\begin{align*}
    \operatorname{MISE}\left\{\widehat{\phi}_j(x),\phi_j(x)\right\} = \mathcal{O}\left(n^{-\frac{4}{4+d}}\right).
\end{align*}
\end{proposition}
% \begin{proof}
% As in appendix \ref{glob_proof}.
% \end{proof}

The proof of Proposition \ref{proposition:conv_comp} can be found in the supplementary material A.1. The main idea is to write the difference between the estimator and the true Shapley curve in terms of a weighted sum,
\begin{align}
\label{eq:weightedsum}
    \widehat{\phi}_j(x)-\phi_j(x) &= \sum_{s \subseteq N} \omega_{j,s} \left\{\widehat{m}_s(x_s)-m_s(x_s)\right\},
\end{align}
where the weights are defined as
\begin{align}
    \omega_{j,s}&=\text{sgn}\{j \in s\}\frac{1}{d}\binom{d-1}{|s| - \mathbf{I}\{j \in s\}  }^{-1}
\end{align}
for all $s\in\mathcal{S}$.
We show that the leading term in the MISE of the component-based estimator depends on the MISE of the local linear estimator for the full model, and since the corresponding weight is defined as $\omega_{j,m}=\frac{1}{d}$,
\begin{align*}
    \operatorname{MISE}\left\{\widehat{\phi}_j(x),\phi_j(x)\right\} = \frac{1}{d^2}\operatorname{MISE}\left\{\widehat{m}(x),m(x)\right\}+{\scriptstyle\mathcal{O}}(n^{-\frac{4}{4+d}}).
\end{align*}
Ultimately, the convergence rate is determined by the slowest rate of the components, i.e. the convergence rate of the full model, which is known to be $\mathcal{O}(n^{-\frac{4}{4+d}})$.
Since $\phi_j(x)$ also belongs to $\mathcal{M}_d$, the class of twice continuously differentiable functions, $\widehat{\phi}_j(x)$ is a minimax-optimal estimator for $\phi_j(x)$ by \citet{stone1982optimal}.

The following theorem establishes the point-wise asymptotic normality of the component-based estimator and the corresponding proof is given in the supplementary material A.2. 

\begin{theorem}
\label{theorem:comp_normal}
Let the conditions of Proposition \ref{proposition:conv_comp} hold and let $h_m\sim n^{-\frac{1}{4+d}}$ denote the optimal bandwidth of the full model. Then we have, for a point $x$ in the interior of $\mathcal{X}$, as $n$ goes to infinity,
\begin{align*}
    \sqrt{nh^d_m}\left\{\widehat{\phi}_j(x)-\phi_j(x)\right\}=\sqrt{nh^d_m}\frac{1}{d}\left\{\widehat{m}(x)-m(x)\right\}+{\scriptstyle\mathcal{O}}_p(1)\stackrel{\mathcal{L}}{\rightarrow}N\left(B(x),V(x)\right),
\end{align*}
where the asymptotic bias and the asymptotic variance are given as
\begin{align*}
    B(x)=\frac{1}{d} \frac{\mu_2(k)}{2} \sum^d_{j=1} \frac{\partial^2 m(x)}{\partial^2 x_j}\ \ \text{and}\  \ V(x)=\frac{1}{d^2} ||k||^2_2 \frac{\sigma^2(x)}{f(x)},
\end{align*}
respectively and $||k||^2_2=\int k^2(s)ds$ denotes the squared $L_2$ norm of $k$.
\end{theorem}
       
Interestingly, the asymptotic distribution of the component-based estimator is the same for all variables, $j=1,\ldots,d$, as neither bias nor variance are variable-specific. The estimators for the bias and the variance, $\widehat{B}(x)$ and $\widehat{V}(x)$ respectively, can be obtained via plug-in estimates. This allows us to construct asymptotically valid confidence intervals around the estimated Shapley curves.

%Motivate the reason for using bootstrap at all
%\textcolor{red}{klarer formulieren:}
%Unfortunately, the established convergence rate is known to result in %under-coverage for finite samples, due to its slow convergence %\citep{song2012bootstrap}. For this reason, we use the bootstrap, as %described in the supplementary material. In the nonparametric %literature, it is well known that bootstrap sampling yields improved %finite sample coverage rather than directly estimating the confidence %intervals relying on the asymptotic normal distribution %\citep{hardle1991bootstrap, hardle1993comparing, hardle2015tie}. 

It is well known that bootstrap sampling, particularly the \textit{wild} bootstrap \citep{mammen1992bootstrap}, yields improved finite sample coverage compared to the direct estimation of the confidence intervals relying on the asymptotic normal distribution \citep{hardle1991bootstrap, hardle1993comparing}. Bootstrap methods in the Shapley context were previously studied by \citet{covert2021improving} and \citet{johnsen2021inferring}. However, the former are interested in the quantification of uncertainty caused by computational approximation methods while our focus is on the estimation uncertainty. The latter also focus on estimation uncertainty but do not provide any theoretical justification for their bootstrap procedure. We propose Algorithm 1, a tailored wild bootstrap procedure to construct asymptotically valid confidence intervals around the component-based estimator.
\begin{algorithm}
\caption{Wild bootstrap procedure for the component-based estimator}
\begin{algorithmic}
\State \textbf{1: Estimate $\widehat{m}_s(x_s)$ on $(X_i,Y_i)_{i=1}^{n}$, with the optimal bandwidth $h_s$ for $s\in\mathcal{S}$ and calculate $\widehat{\phi}_j(x)$.}
\State \textbf{2: Estimate $\widehat{m}_{s,g}(x_s)$ on $(X_i,Y_i)_{i=1}^{n}$, with bandwidth $g_s$ such that $\frac{h_s}{g_s} \rightarrow 0$ as $n \rightarrow \infty$ for all $s\in\mathcal{S}$ and calculate $\widehat{\phi}_{j,g}(x)$.}

\State \textbf{3: Bootstrap sampling}
\begin{enumerate}[label=(\alph*)]
    \item Calculate the bootstrap residuals by using wild bootstrap, such that $\varepsilon^{*}_{i,s} = \widehat{\varepsilon}_{i,s} \cdot V_i$, where $\widehat{\varepsilon}_{i,s} = Y_i - \widehat{m}_s(X_{i,s})$ for all $s\in\mathcal{S}$. As introduced in \citet{mammen1993bootstrap}, the random variable $V_i$ is $-(\sqrt{5}-1) / 2$ with probability $(\sqrt{5}+1) /(2 \sqrt{5})$ and $(\sqrt{5}+1) / 2$ with probability $(\sqrt{5}-$ 1) $/(2 \sqrt{5})$. 
    
    \item Construct $Y_{i,s}^{*}= \widehat{m}_{s,g}(X_{i,s}) + \varepsilon_{i,s}^{*}$ for $i=1,\ldots,n$ and for all $s\in\mathcal{S}$.

    \item Estimate $\widehat{m}^{*}_{s}(X_s)$ based on the bootstrap version $(X_i,Y_{i,s}^{*})_{i=1}^{n}$ with bandwidths $h_s$ and calculate $\widehat{\phi}^{*}_{j,b}(x)$.
    
\end{enumerate}
\State \textbf{4: Iteration}
\State Repeat Step 3(a) - 3(c) for $b=1,\ldots,B$ bootstrap iterations.  
\State \textbf{5: Construct confidence intervals}
\State Construct confidence intervals $CI\left\{\phi_j (x)\right\} = \left\{\widehat{\phi}_j(x) + q_{\frac{\alpha}{2}}, \widehat{\phi}_j(x) + q_{1 - \frac{\alpha}{2}}\right\}$, where $\alpha$ is the significance level and $q_{\frac{\alpha}{2}}$ and $q_{1 - \frac{\alpha}{2}}$ are the empirical quantiles of the bootstrap distribution of $\widehat{\phi}^{*}_{j}(x) - \widehat{\phi}_{j,g}(x)=\sum_{s \subseteq N} \omega_{j,s} \left\{   \widehat{m}^{*}_{s}(x_s) - \widehat{m}_{s,g}(x_s)     \right\} $.
\end{algorithmic}
\end{algorithm}

The novelty of our proposed bootstrap procedure is the creation of subset-specific bootstrap observations $(X_i,Y_{i,s}^{*})_{i=1}^{n}$, and the inclusion of the lower-order components. The procedure is presented in Algorithm 1. By incorporating the subset-specific bootstrap data, the variance of the bootstrap version better mimics the variance of the estimator in finite samples, as shown in Section A.4 in the supplementary material. Even though the lower-order components are irrelevant to the asymptotic distribution, they do matter in finite samples. Further, it is crucial to choose the bandwidth $g_s$ such that it oversmooths the data to correctly adjust the bias in the bootstrap version of the Shapley curves \citep{hardle1991bootstrap}. The following proposition presents the consistency of our bootstrap procedure. For this result, we have to assume that the third moment of the error term is bounded.

\begin{assumption}\label{assum:bs}
Assume that the conditional variance $\sigma^2(x)$ is twice continuously differentiable and $\sup_{x} {\sf E}(\varepsilon^3 \lvert X=x) < \infty$.  
\end{assumption}

\begin{proposition}
\label{proposition:bs}
Let Assumptions \ref{assum:density}-\ref{assum:bs} hold and let $P^{Y \mid X}$ denote the conditional distribution and $P^{*}$ denote the bootstrap distribution. Then we have, for a point $x$ in the interior of $\mathcal{X}$ and $z \in \mathbb{R}$, as $n$ goes to infinity
\begin{align*}
   \bigg \lvert P^{Y \mid X}\left[\sqrt{n h_m^d}\left\{\widehat{\phi}_j(x)-\phi_j(x)\right\}<z\right]-  P^{*}\left[\sqrt{nh^d_m}\left\{\widehat{\phi}^{*}_j(x)-\widehat{\phi}_{g,j}(x)\right\} <z \right] \bigg| {\rightarrow} 0.
\end{align*}
\end{proposition}

The underlying function class in Proposition \ref{proposition:conv_comp} and Theorem \ref{theorem:comp_normal} is quite large, leading to a prohibitively slow convergence rate in a setting with high-dimensional covariates due to the curse of dimensionality. However, we want to emphasize that it is possible to obtain convergence rate results for function classes more suited for such higher-dimensional settings. A relevant example is the function class considered in \citet{schmidt2020nonparametric}, containing elements that are compositions of lower-dimensional functions with dimensions up to $d^*$, which can be much smaller than the original covariate dimension $d$. Deep neural networks with ReLU activation functions can estimate such functions with the rate $n^{-2/(4+d^*)}$, assuming that the underlying lower-dimensional functions are twice continuously differentiable. In the case of independent covariates, this favorable convergence rate can also be achieved for the estimation of Shapley curves when deep learning is used in the estimation of the components in (\ref{eq:shapley_est}). The result of \citet{schmidt2020nonparametric} can directly be applied to Proposition \ref{proposition:conv_comp}. As a downside, while we know the rate of convergence, asymptotic normality results as in Theorem \ref{theorem:comp_normal} are much harder to obtain and are still an open research problem.

%3. On WLS
\begin{remark}
The component-based estimator for the Shapley curves, $\widehat{\phi}_j(x)$, requires the estimation of $2^d$ conditional mean functions, $m_s(x_s)$, for all $s\in\mathcal{S}$. This task is particularly cumbersome when $d$ is large and the chosen estimator is computationally intensive. The KernelSHAP approximation mitigates this issue by subsampling the subsets $s\in\mathcal{S}$. This subsampling idea can be incorporated into a constrained weighted least-squares problem, whose solution offers an approximation to the usual analytical formula. Recently, \citet{jethani2022fastshap} extended this idea to FastSHAP, an online method for estimating Shapley values based on stochastic gradient descent. We refer to \citet{chen2023algorithms} for a comprehensive overview of approximation methods. Since our focus is to carefully derive the main asymptotic properties of the introduced estimators, we do not provide new solutions to computational issues. However, it is straightforward to incorporate these approximation methods into our statistical analysis. The reason is that as the number of sampled subsets goes to infinity reasonably fast, the approximation error is of smaller order compared with the estimation error. 
% \citet{williamson2020efficient} provide an approximation technique which is based on subsampling the subsets $s\in\mathcal{S}$. This subsampling idea is incorporated into a constrained weighted least-squares problem. The authors derive a closed-form solution, which drastically reduces the computational cost. In principle, we can directly incorporate both of our estimation approaches into their closed-form solution and obtain approximated Shapley curves. 
% Further, we know that the approximated curves converge to the population counterpart with the parametric rate as the number of subsamples goes to infinity. Although the inclusion of this approximation is straightforward for our asymptotic results, it is of no relevance as long as the subsample size approaches infinity.  
\end{remark}

\subsection{Integration-Based Approach}
\label{integ_chapter}

In contrast to the component-based method, the integration-based approach requires the estimation of only one regression function, namely that of the full model. This pilot estimator is obtained by local linear estimation and is thus identical to the estimator for the full model in the component-based approach, $\widehat{m}(x)$. To estimate the regression functions for the subsets $s\in\mathcal{S}$, the variables not contained in $s$ are integrated out using the pilot estimator and an estimate for the conditional density,
\begin{align}
    \widetilde{m}_s(x_s)=\int_{\mathbb{R}^{d-|s|}} \widehat{m}(x)\widehat{f}_{X_{-s}|X_s}(x_{-s}|x_s)dx_{-s}.
    \label{eq:dependent}
\end{align}
In the case of mean-independent regressors, there is no need to estimate conditional densities. In this case, a simplified estimator can be used, which averages over the observations,
\begin{align}
    \widetilde{m}_s(x_s)=\frac{1}{n}\sum_{i=1}^{n}\widehat{m}(X_s=x_s,X_{-s,i}).
    \label{eq:independent}
\end{align}
This estimator is well-known in the literature as the marginal integration estimator for additive models \citep{linton1995kernel,tjostheim1994nonparametric} and is discussed in a more general setting by \citet{fan1998direct}. However, the assumption of mean-independent regressors is not plausible in general, and the estimator described in (\ref{eq:independent}) will lead to inconsistent estimates for the true component, $m_s(x_s)$. In this setting, the estimation of the conditional densities is a necessity and one needs to use the estimator (\ref{eq:dependent}).

The estimation of the conditional density is somewhat cumbersome in practice. For example, \citet{aas2021explaining} use a similar definition of the conditional mean as what we call \textit{integration-based approach} (see equation \ref{eq:component}). However, they do not focus on the resulting asymptotic properties but rather provide practical approaches to tackle the estimation problem of the conditional density. These approaches range from assuming variable independence, a Gaussian distribution, or vine copula structures. Going further, \citet{chen2023algorithms} provide a widespread summary of approximation techniques to the same estimation problem.

To study the asymptotic behavior of the integration-based estimator for Shapley curves, we assume the simplified setting of known conditional densities. Thus, we consider the following estimator for the component function of subset $s$,
\begin{align}
\label{eq:integ}
    \widetilde{m}_s(x_s)=\int\widehat{m}(x)f_{X_{-s}|X_s}(x_{-s}|x_s)dx_{-s}.
\end{align}
In analogy to the component-based estimated Shapley curve (\ref{eq:shapley_est}), we obtain the integration-based estimated Shapley curve  $\widetilde{\phi}_j(x)$ as a plug-in estimate using $\widetilde{m}_s(x_s)$.

The following theorem shows the global convergence rate and the asymptotic distribution of the integration-based estimator.
\begin{theorem}
\label{theorem:conv_integ}
Under Assumptions \ref{assum:density}, \ref{assum:function} and \ref{assum:kernel}, let $\widetilde{\phi}_j(x)$ be the integration-based estimator with known density and a pilot estimator based on local linear estimation with bandwidth $h_m\sim n^{-\frac{1}{4+d}}$. Then we have as $n$ goes to infinity,
\begin{align}
    \operatorname{MISE}\left\{\widetilde{\phi}_j(x),\phi_j(x)\right\} = \mathcal{O}\left(n^{-\frac{4}{4+d}}\right). \label{eq:conv_integ}
\end{align}
Further, we have, for a point $x$ in the interior of $\mathcal{X}$, as $n$ goes to infinity the asymptotic distribution of the integration-based estimator,
\begin{align*}
    \sqrt{nh^d_m}\left\{\widetilde{\phi}_j(x)-\phi_j(x)\right\} =  \sqrt{nh^d_m}\sum_{s \subseteq N} w_{j,s}  \left\{  \widetilde{m}_s(x_s) - m_s(x_s) \right\}\stackrel{\mathcal{L}}{\rightarrow}N\left(B_{int}(x),V(x)\right),
\end{align*}
where the asymptotic bias term is 
\begin{align}
B_{int}(x)= \frac{\mu_2(k)}{2} \sum_{s \subseteq N} w_{j,s} \left\{ \sum^d_{j=1} \int_{X_{-s}} \frac{\partial^2 m(x)}{\partial^2 x_j} f_{X_{-s}|X_s}(x_{-s}|x_s)dx_{-s}\right\}
\label{eq:bias}
\end{align}
and the asymptotic variance term is
\begin{align*}
    V(x)=\frac{1}{d^2} ||k||^2_2 \frac{\sigma^2(x)}{f(x)}.
\end{align*}
\end{theorem}

The proof of Theorem \ref{theorem:conv_integ} is given in the supplementary material A.5. Notice that the rate is identical to the convergence rate of the component-based approach and it is also optimal in the minimax sense. The reason is that, once again, the (pilot) estimator $\widehat{m}(x)$ determines the convergence rate of $\widetilde{\phi}_j(x)$.

The asymptotic variance is identical to that of the component-based approach. The difference in the asymptotic distribution is solely due to the bias. Compared with the component-based approach, the bias term defined in (\ref{eq:bias}) is now a sum over $2^d$ elements, instead of a single term. As a consequence, the bias is inflated. This fact becomes clear when looking at the bias and the variance of the estimated components. Following \citet{linton1995kernel}, we get the following expression for the bias and the variance of $\widetilde{m}_s(x_s)$,
\begin{align*}
    \operatorname{Bias}\left\{\widetilde{m}_s(x_s)\right\}&=\frac{1}{2}h_m^2\mu_2(k)\int\sum_{j=1}^{d}\frac{\partial^2 m(x_s)}{\partial x_{j}^2}f_{X_{-s}|X_s}(x_{-s}|x_s)dx_{-s}+{\scriptstyle\mathcal{O}}(h_m^2),\\
    \operatorname{Var}\left\{\widetilde{m}_s(x_s)\right\}&=\frac{1}{nh_m^{|s|}}||k||^2_2\int \frac{\sigma^2(x)f^2_{X_{-s}|X_s}(x_{-s}|x_s)}{f(x)}dx_{-s}+{\scriptstyle\mathcal{O}}\left(\frac{1}{nh_m^{|s|}}\right).
\end{align*}
The bandwidth of the pilot estimator is chosen as $h_m\sim n^{-\frac{1}{4+d}}$, which balances the squared bias and variance only for the component associated with the full model. However, all components are based on this pilot estimator and thus rely on the same bandwidth. For all other subsets $s$, we have $|s|<d$ which leads to oversmoothing,
\begin{align*}
    \operatorname{Bias}^2\left\{\widetilde{m}_s(x_s)\right\}&=\mathcal{O}(n^{-\frac{4}{4+d}})\\
    \operatorname{Var}\left\{\widetilde{m}_s(x_s)\right\}&=\mathcal{O}(n^{-\frac{(4+d-|s|)}{4+d}}).
\end{align*}
As a consequence, the bias of the lower-dimensional components does not vanish in the bias of the integration-based estimator, as it is of the same order for all the components, namely $\mathcal{O}(n^{\frac{2}{4+d}})$. While the order of the bias term in the integration-based estimator is still identical to that of the component-based approach, it might be substantially larger in finite samples.
However, the advantage of the integration-based curve is that the pilot estimator only needs to be estimated once. Essentially, this constitutes a trade-off between computational complexity and accuracy in the estimation.

\begin{remark}
While the focus in this Section lies on the local linear estimator and the class of twice-continuously differentiable functions, the bias problem of the integration-based approach presumably also arises in other contexts. This is due to the pilot estimator, which typically involves the selection of hyperparameters governing the bias-variance trade-off. For random forests, this could be the depth of the individual trees, for neural networks, the number of layers, and number of nodes. There is no reason to believe that the optimal choice of hyperparameters for the pilot estimator, which is based on the set of all regressors, is optimal for the estimation of the components associated with the lower-dimensional subsets. These components are less complex than the full model, which the integration-based approach cannot accommodate, while the component-based approach can.

\end{remark}

\begin{remark}
The result of Theorem \ref{theorem:conv_integ} relies on the knowledge of the true conditional densities for the integration-based estimation of the components. In practice, these densities have to be estimated, introducing another term in the asymptotic expression. For example, a Gaussian distribution is a reasonable assumption \citep{aas2021explaining1}, leading to a smaller order term of rate $\sqrt{n}$. Besides, practitioners could assume a hierarchical dependence structure of the variables and estimate the conditional density via parametric or nonparametric vine copula approaches \citep{aas2021explaining}. For a kernel-density estimator, this additional estimation error is of order $\mathcal{O}_p(n^{-\frac{2}{4+d}})$. This is a further disadvantage of the integration-based method of estimating Shapley curves compared with the component-based approach.
\end{remark}

% motivation of additive model:
The previous theorems are affected by the curse of dimensionality. Additive models are known to bypass this problem by imposing structure on the true process. They are renowned for offering a good balance between model flexibility and ease of interpretation. For example, \citet{scornet2015consistency} assume an additive structure on the true process for random forest estimation. Under the assumption of additivity and independence of the explanatory variables, we know that the Shapley curve simplifies to the corresponding partial dependence function, as explained in Example 1.

\begin{assumption}\label{assum:additive}
Assume the regression function $m(x)$ follows an additive structure, s.t. $m(x)=\sum^d_{j=1} g_j(x_j)$ with ${\sf E}\{g_j(x_j)\}=0$ for $j=1,\ldots,d$ and the explanatory variables are independent.
\end{assumption}
It follows directly by \citet{stone1985additive} that the partial dependence function, and thus the corresponding Shapley curve, can be estimated with a one-dimensional rate, see Corollary \ref{col:additive}. The estimation of the partial dependence function is sufficient to obtain an estimator for $\phi_j(x)$. 
%This can be done consistently via marginal integration \citep{linton1995kernel} or backfitting \citep{mammen1999existence}. 
%Further insights into the interpretation of smooth backfitting and marginal integration for additive models is given in \citet{nielsen1998optimization}.
The bias and the variance of the estimated Shapley curves follow by the established asymptotic results for the estimator of the partial dependence function for variable $j$ \citep{linton1995kernel, nielsen1998optimization}. 
Note that recent results relying on lower-order terms in a functional decomposition of the regression function such as \citet{hiabu2023unifying}, can directly be applied to the Shapely curves. In that case, the order of approximation determines the convergence rate.

\begin{corollary}\label{col:additive}
Let the partial dependence function $\widehat{g}_j(x_j)$ be an estimator for $\phi_j(x)$ obtained by marginal integration or backfitting. Under Assumptions \ref{assum:density}, \ref{assum:function}, \ref{assum:kernel} and \ref{assum:additive} we have that
\begin{align*}
    \operatorname{MISE}\left\{\widehat{g}_j(x),\phi_j(x)\right\} = \mathcal{O}\left(n^{-\frac{4}{5}}\right).
\end{align*}
\end{corollary}

\section{Numerical Studies}
\label{section:numerical}
%1. Einleitung
In this section, we conduct simulation studies to validate the previous asymptotic results. First, we empirically demonstrate the consistent estimation of both the component and integration-based estimation techniques. Next, we use the proposed bootstrap approach to show empirical coverage.
%2. DGP

Let the regressors be zero mean Gaussian with variance $\sigma^2=4$ and correlation $\rho$, which is either $0$ or $0.8$ for different setups.
The corresponding density functions are assumed to be known.
The first data-generating process (DGP) is an additive model and the second one includes interactions:
\begin{align*}
&\textbf{\text{DGP 1: Additive}}\ \ \ \ \ \ m(x) = -\text{sin}(2x_1) + \text{cos}(2x_2) + x_3 \\
&\textbf{\text{DGP 2: Interactive}}\ \ \ m(x) = -\text{sin}(2x_1) + \text{cos}(3x_2) + 0.5x_3 + 2\text{cos}(x_1)\text{sin}(2x_2). 
\end{align*}
The error terms follow $\varepsilon \sim N(0,1)$. To investigate for robustness, we include a variation for $\varepsilon \sim t(5)$, which reduces the signal-to-noise ratio.
%3) LL, h, b
The local linear estimator is used to obtain component-based and integration-based Shapley curves for all three variables. The bandwidths $h_1,\ldots,h_d$ differ in each direction and are chosen via leave-one-out cross-validation in each $j$-direction. We use the second-order Gaussian kernel.
%4) On MISE 
To evaluate the global performance of both estimators, we calculate the MISE for DGP 1 and DGP 2 based on $6000$ Monte Carlo replications of the same experiment. The estimated as well as the population-level Shapley curve are illustrated as heatmaps in Figure S1 in the supplementary material, together with the squared error.
%\begin{align}
%\label{mise_ref}
%     \operatorname{MISE}\left\{\widehat{\phi}_j(x),\phi_j(x)\right\} = {\sf E} \left[  \int_{\mathbb{R}^d} \left\{ \widehat{\phi}_j(x) - \phi_j(x)  \right\}^2 dx \right].
%\end{align}
%5) MC runs and n

The simulation results are displayed in Table \ref{tab1} for both DGPs and Gaussian error terms. Recall that $\widetilde{\phi}_j(x)$ and $\widehat{\phi}_j(x)$ are denoted as the estimated integration-based Shapley curve and the estimated component-based Shapley curve for variable $j$, respectively. First, we observe that the MISE is shrinking for both estimators as the sample size increases. This result empirically confirms Proposition \ref{proposition:conv_comp} and the first part of Theorem \ref{theorem:conv_integ}.
Second, the component-based approach (almost) always results in a smaller MISE than the integration-based approach. This aligns with the asymptotic bias and asymptotic variance comparison of both estimators in Chapter \ref{section:estimation}. Since we oversmooth within the integration-based estimator, the bias of $\widetilde{\phi}_j$ accumulates over each component. In contrast, the bias of $\widehat{\phi}_j$ is only determined by a single component, namely the one associated with the full $d$-dimensional model. 
Since the asymptotic variance of $\widetilde{\phi}_j(x)$ and $\widehat{\phi}_j(x)$ is equal, the bias is the crucial part for the better performance of the component-based estimator. Further, Table \ref{tab1} underlines that the bias term is more prominent the more complex the true process is.

%on x3
The Shapley curves for the third variable result in better performance when the integration-based estimation is applied. This observation might seem counterintuitive at first. As the pilot estimator $\widehat{m}(x)$ oversmooths in each dimension, the bandwidth of the third variable, $h_3$, is tuned to take the whole support of $X_3$. Computationally, this happens in every iteration of our Monte Carlo simulation. As a consequence, a linear model is fitted in the $X_3$ direction, which takes the whole support for estimation. This reduces the variance as more observations are available for the fit. In contrast, the component-based estimation is not able to tune $h_3$ such that it captures the whole support in $\widehat{m}_s(x_s)$, for each subset that contains the third variable.

\begin{table}[H]
\spacingset{1.2}
\caption{MISE of Shapley Curves for component-based estimator $\widehat{\phi}_j$ and integration-based estimator $\widetilde{\phi}_j$ for each variable. The additive and interactive DGP is simulated with variable correlations of $\rho=0$ and $\rho=0.8$. The error terms follow $\varepsilon \sim N(0, 1)$. }\label{tab1}
\centering
%\hspace*{-1.1cm}
\setlength{\tabcolsep}{8pt}
\begin{tabular}{ r r r r r r r r r }
 DGP & $\rho$ & $n$ & $\widehat{\phi}_1$ & $\widetilde{\phi}_1$ & $\widehat{\phi}_2$ & $\widetilde{\phi}_2$ &$\widehat{\phi}_3$ & $\widetilde{\phi}_3$  \\
\hline
\multirow{8}{*}{Additive} & \multirow{4}{*}{$0$} & 300 & \bftab8.84 & 14.79 & \bftab8.53 & 8.72 & 3.05 & \bftab{0.34} \\ 
  & & 500 & \bftab{4.86} & 5.37 & \bftab{5.47} & 6.30 & 1.91 & \bftab{0.22} \\ 
  & & 1000 & \bftab{3.04} & 3.29 & \bftab{3.48} & 3.87 & 1.09 & \bftab{0.11} \\ 
  & & 2000 & \bftab{1.83} & 1.91 & \bftab{2.13} & 2.43 & 0.66 & \bftab{0.07} \\ 
\cline{2-9} 
& \multirow{4}{*}{$0.8$} & 300 & \bftab{7.29} & 7.81 & \bftab{7.99} & 8.87 & 2.69 & \bftab{1.27} \\ 
 & & 500 & \bftab{5.09} & 5.41 & \bftab{5.65} & 6.05 & 1.86 & \bftab{0.83} \\ 
 & & 1000 & \bftab{3.37} & 3.38 & \bftab{3.46} & 3.79 & 1.27 & \bftab{0.52} \\ 
 & & 2000 & 2.24 & \bftab{2.10} & \bftab{2.22} & 2.37 & 0.87 & \bftab{0.34} \\ 
\hline
\multirow{8}{*}{Interactive} & \multirow{4}{*}{$0$} & 300 &  \bftab{9.31} & 12.76 & \bftab{11.45} & 14.15 & 1.74 & \bftab{0.40} \\ 
 & & 500 & \bftab{5.68} & 7.51 & \bftab{6.74} & 7.86 & 0.87 & \bftab{0.23} \\ 
 & & 1000 & \bftab{3.20} & 4.20 & \bftab{3.92} & 4.56 & 0.51 & \bftab{0.13} \\ 
 & & 2000 & \bftab{1.90} & 2.45 & \bftab{2.22} & 2.63 & 0.30 & \bftab{0.08} \\ 
\cline{2-9} 
& \multirow{4}{*}{$0.8$} & 300 & \bftab{10.85} & 12.04 & \bftab{11.89} & 14.34 & 3.16 & \bftab{1.81} \\ 
 & & 500 & \bftab{7.59} & 7.82 & \bftab{7.98} & 9.79 & 2.20 & \bftab{1.27}  \\ 
 & & 1000 & 4.99 & \bftab{4.70} & \bftab{4.79} & 5.94 & 1.50 & \bftab{0.79} \\ 
 & & 2000 & 3.27 & \bftab{2.86} & \bftab{3.00} & 3.57 & 1.05 & \bftab{0.49} \\ 
\hline
\end{tabular}
\end{table}

%On additive model
Note that we do not assume knowledge of the true process during the estimation of Shapley curves. If we conduct the estimation knowing that the first DGP is of additive structure, we can use backfitting or marginal integration. In this case, the optimal one-dimensional bandwidth for dimension $j$ would be used, instead of the full model bandwidth. Following Corollary \ref{col:additive} a faster decrease of the MISE will result since the convergence is one-dimensional.
%Since we conduct the estimation without knowledge of the true process, we get the same bias story as for the non-linear DGP.
%robustness: correlation and t(5)
Further, we conclude that the Shapley curves are robust to a lower signal-to-noise ratio as we obtain reasonable results for $\varepsilon \sim t(5)$ in Table S1 the supplementary material.
Both tables show that including dependence between the variables in the form of correlation introduces indirect effects as described in Example 1. This leads to an increase in the MISE for (almost) all considered sample sizes. We have included further simulation results for higher dimensional settings in Table S3 in the supplementary material. These results are accompanied by computational run times, illustrating the practical feasibility. 

In the subsequent paragraphs, we are going to investigate the empirical coverage probability of the confidence intervals for the component-based Shapley curves. Due to the better convergence performance as well as the smaller asymptotic bias, we set our focus merely on the component-based Shapley curves. The proposed wild bootstrap procedure from Section \ref{comp_chapter} is applied. 
Let the true process be a non-linear function of $X_1$ and $X_2$:
\begin{align}
\label{bs_dgp}
\textbf{\text{DGP 3:}}\ \ \ \ \ \ m(x) = -\text{sin}(2x_1) + 0.1 x_2 + 2\text{cos}(x_1)\text{sin}(x_2). 
\end{align}

The estimated coverage probabilities for $M=1000$ Monte Carlo replications are reported in Table \ref{cov_n_bs} for the component-based estimation, with Gaussian errors $\varepsilon \sim N(0, 1)$ and $t$-distributed error terms. It demonstrates that our bootstrap procedure works well in finite samples since we obtain the desired coverage ratios. Further, the estimation is robust to increased noise in the true process.
%On coverage main BS table
As expected, the local linear estimation is accompanied by a bias term which leads to a slight under-coverage in our simulation setup. Further. Figure S2 in the supplementary material illustrates the estimated curve, the population curve as well as the bootstrap confidence intervals.

\begin{table}[H]
\spacingset{1.2}
\caption{Estimated coverage probability for component-based Shapley curves for $\varepsilon \sim t(5)$ and $\varepsilon \sim N(0, 1)$ and significance levels $\alpha=0.15, 0.1, 0.05$. }\label{cov_n_bs}
\centering
\begin{tabular}{rrrrrrrrrrrrr}
  \hline
     & \multicolumn{6}{c}{Variable 1} & \multicolumn{6}{c}{Variable 2}\\
  \hline
  $n$ & \multicolumn{3}{c}{$N(0, 1)$} & \multicolumn{3}{c}{$t(5)$} & \multicolumn{3}{c}{$N(0, 1)$} & \multicolumn{3}{c}{$t(5)$}\\\cmidrule{2-13}
 & 0.15 & 0.1 & 0.05 & 0.15 & 0.1 & 0.05 & 0.15 & 0.1 & 0.05 & 0.15 & 0.1 & 0.05 \\ 
  \hline
100 & 0.79 & 0.85 & 0.90 & 0.79 & 0.86 & 0.92 & 0.74 & 0.79 & 0.85 & 0.73 & 0.80 & 0.86  \\ 
  250 & 0.82& 0.87&  0.93&  0.83&  0.87&  0.92&  0.76&  0.81&  0.89&  0.76&  0.81&  0.87 \\ 
  500 & 0.82& 0.86&  0.93&  0.83&  0.87&  0.92&  0.77&  0.82&  0.90&  0.80&  0.85&  0.91 \\ 
  1000 & 0.82&  0.87&  0.93&  0.83&  0.87&  0.91&  0.79&  0.84&  0.91&  0.81&  0.86&  0.92\\ 
  2000 & 0.82&  0.87&  0.93&  0.82&  0.87&  0.94&  0.79&  0.85&  0.91&  0.82&  0.89&  0.94 \\ 
  4000 & 0.85& 0.90&  0.95&  0.82&  0.88&  0.94&  0.81&  0.86&  0.93&  0.84&  0.90&  0.95 \\ 
   \hline
\end{tabular}
\end{table}

\section{Empirical Application}
\label{section:application}
%Einleitung
This empirical application illustrates our previous results in a real-world situation of vehicle price setting. Vehicle manufacturers typically rely on price-setting approaches that involve teardown data, surveying, or indirect cost multiplier adjustments for calculating a markup. In contrast, we apply a data-driven approach as a potential alternative. The sample consists of extensive vehicle price data for the U.S. provided by \citet{moawad2021explainable} and collected by the Argonne National Laboratory. Our data set includes the $20$ most important variables identified by the latter authors out of a larger pool of characteristics for the prediction of $38435$ vehicle prices ranging from the year 2001 to the year 2020.
However, our motivation differs in the sense that we are interested in the visualization of Shapley curves over the whole support of the variables at hand, instead of being restricted to have a variable importance measure only for the observations in the sample.
Our goal is not to obtain the most precise prediction of vehicle prices, but rather to provide a nuanced interpretation approach.

For this matter, we make use of the three most important variables of the data set, which are horsepower, vehicle weight (in pounds), and vehicle length (in inches). In principle, the dimension could be larger but it would not contribute to an interpretable empirical application. The data is divided into three time intervals, ranging from 2001--2007 (12230), 2008--2013 (11955), and 2014--2020 (14250). The pooled summary statistics can be found in Table S2 in the supplementary material. In the following, the vehicle price is reported in $1000$ USD.

    % Why is it interesting to look at Shapley Curves (Inhaltliche motivation, welche etwas über Autos und Auto-Pricing beinhaltet.
 
    It is of empirical interest to consider Shapley curves as proposed in this work for several reasons. First, we are able to decompose the estimated conditional expectation locally at any point vector of interest. This type of investigation can contribute to an empirical understanding of prices for U.S. vehicle companies. Second, we obtain asymptotically valid confidence intervals around the estimated Shapley curves. This enables the price setter to differentiate between significant and non-significant support sections of the covariates.
    % On estimation and bandwidth selection 
   In this application, we use the component-based Shapley curves as proposed in the previous chapters. The bandwidth choice is motivated by a dual objective, on the one hand, to get a good fit to the data, and on the other hand to get interpretable and smooth curves.

    % 3D plots Shapley curves. 
   To graphically illustrate the Shapley curves in dependence of two variables, we fix the third variable at the median. The surface plot for the first variable, Horsepower (hp), is illustrated in Figure S3 in the supplementary material in dependence of horsepower and vehicle weight for the pooled data from years 2001--2020. This plot allows us to illustrate the interactive contribution on the price prediction between the variables. As we see, very light cars with high horsepower such as sports cars, are resulting in the highest increase in the price prediction. The lighter the car as well as the higher the horsepower, the higher the contribution to the price.

%    \begin{figure}[H] 
%    \begin{minipage}{0.25\textwidth}
%        \includegraphics[scale=0.182]{first_x2.png} 
%      \end{minipage}
%      \begin{minipage}{0.25\textwidth}
%        %\hspace{-2.4cm}
%        \includegraphics[scale=0.182]{second_x2.png} 
%      \end{minipage}
%      \begin{minipage}{0.5\textwidth}
%         % \hspace{-4.8cm}
%        \includegraphics[scale=0.182]{third_x2.png} 
%        \includegraphics[scale=0.182]{full_x2.png} 
%      \end{minipage} 
%    \caption{Estimated Shapley curves for vehicle weight in dependence of weight (in lbs) for a vehicle length of $190$ inches and $190$ horsepower. The time periods are 2001 - 2007 (top left), 2008 - 2013 (top right), 2014 - 2020 (bottom left) and pooled (bottom right). HAUPT}
%    \label{lbs_slice}
%    \end{figure}

\begin{figure}[ht] 
\begin{minipage}{0.32\textwidth}
    \includegraphics[scale=0.23]{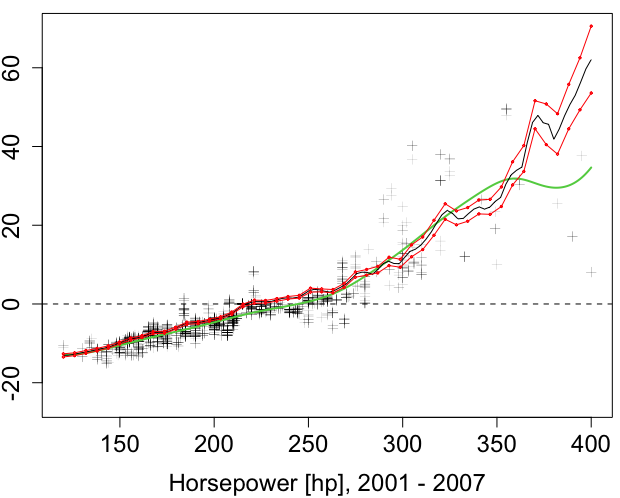} 
  \end{minipage}
  \begin{minipage}{0.32\textwidth}
    \includegraphics[scale=0.23]{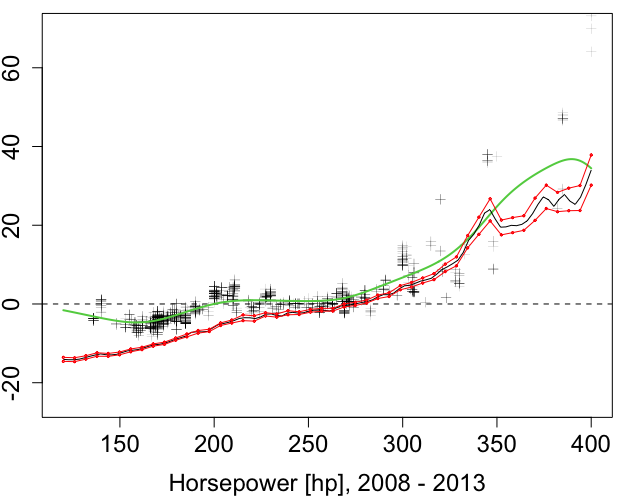} 
  \end{minipage}
  \begin{minipage}{0.32\textwidth}
    \includegraphics[scale=0.23]{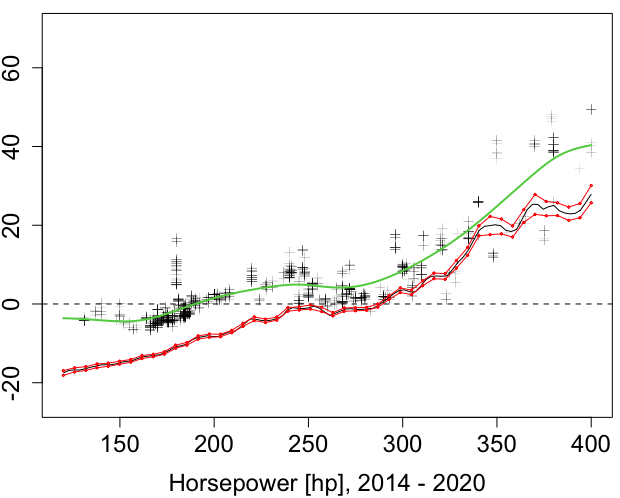} 
  \end{minipage} 
\caption{ Estimated component-based Shapley curves for horsepower in dependence of horsepower (in hp) for a vehicle length of $190$ inches and vehicle weight of $3500$ pounds. The time periods are 2001--2007, 2008--2013, and 2014--2020. Estimated SHAP values (black crosses) and smoothed curve based on these values (green curve).}
\label{hp_slice_shap}
\end{figure}

\begin{figure}[ht] 
\begin{minipage}{0.32\textwidth}
    \includegraphics[scale=0.225]{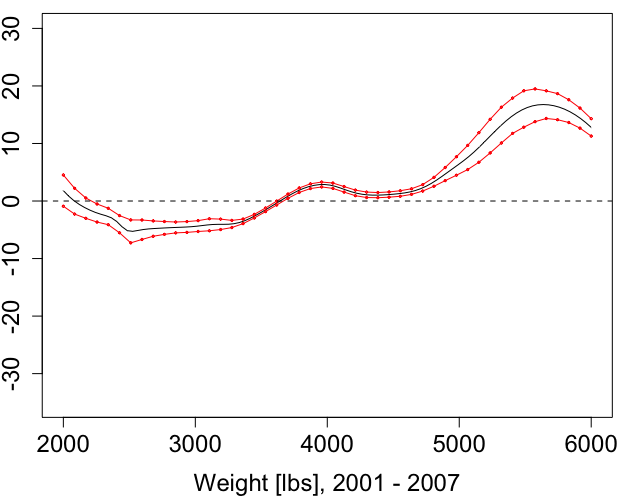} 
  \end{minipage}
  \begin{minipage}{0.32\textwidth}
    \includegraphics[scale=0.225]{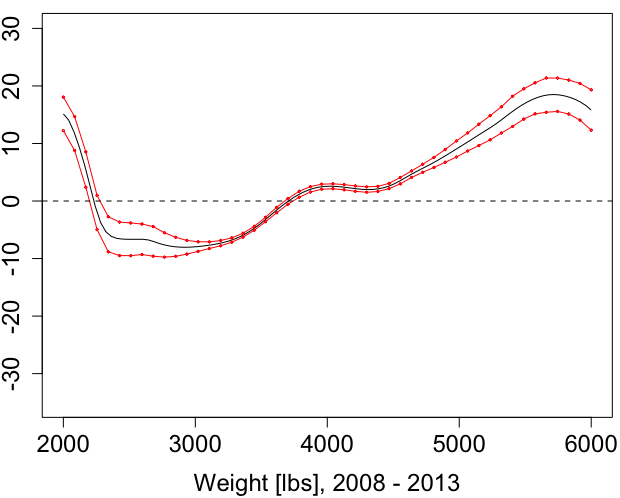} 
  \end{minipage}
  \begin{minipage}{0.32\textwidth}
    \includegraphics[scale=0.225]{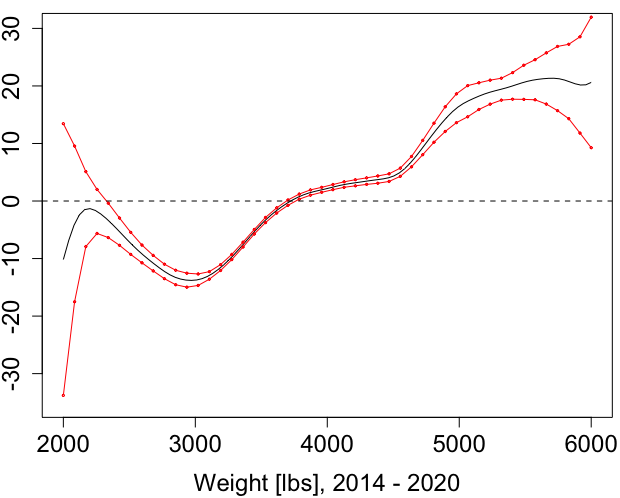} 
  \end{minipage} \\
  \begin{minipage}{0.32\textwidth}
    \includegraphics[scale=0.225]{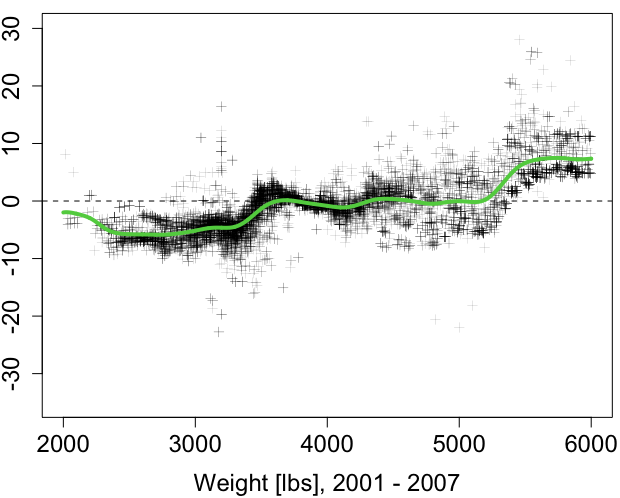} 
  \end{minipage}
  \begin{minipage}{0.32\textwidth}
    \includegraphics[scale=0.225]{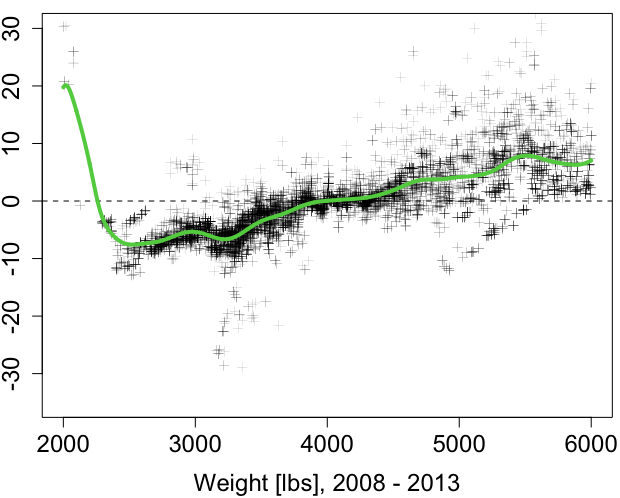} 
  \end{minipage}
  \begin{minipage}{0.32\textwidth}
    \includegraphics[scale=0.225]{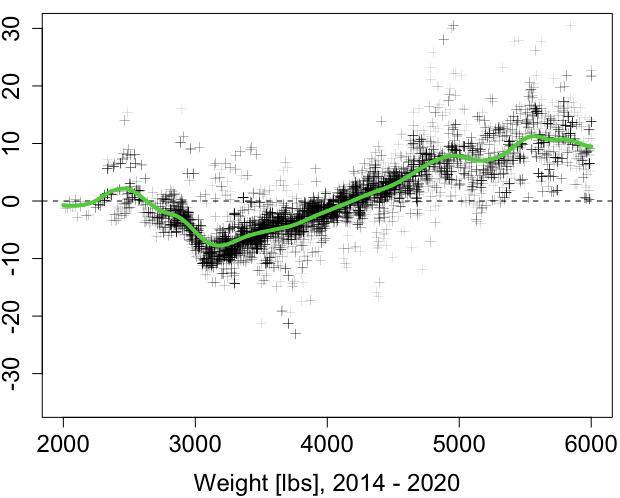} 
  \end{minipage} 
\caption{First row: Estimated component-based Shapley curves for vehicle weight in dependence of weight (in lbs) for a vehicle length of $190$ inches and $190$ horsepower. The time periods are 2001--2007, 2008--2013, and 2014--2020. Second row: Estimated KernelSHAP values (black crosses) and smoothed curve based on these values (green curve).}
\label{lbs_slice}
\end{figure}

%\begin{figure}[ht] 
%\begin{minipage}{0.32\textwidth}
%    \includegraphics[scale=0.23]{first_x2n.png} 
%  \end{minipage}
%  \begin{minipage}{0.32\textwidth}
%    \includegraphics[scale=0.23]{first_x2n.png} 
%  \end{minipage}
%  \begin{minipage}{0.32\textwidth}
%    \includegraphics[scale=0.23]{first_x2n.png} 
%  \end{minipage} 
%\caption{TEST PLOT: Estimated Shapley curves for vehicle weight in dependence of weight (in lbs) for a vehicle length of $190$ inches and $190$ horsepower. %The time periods are 2001 - 2007 (top left), 2008 - 2013 (top right), 2014 - 2020 (bottom left) and pooled (bottom right).}
%\label{lbs_slice}
%\end{figure}

%\begin{figure}[ht] 
%\begin{minipage}{0.32\textwidth}
%    \includegraphics[scale=0.23]{first_x3n.png} 
%  \end{minipage}
%  \begin{minipage}{0.32\textwidth}
%    \includegraphics[scale=0.23]{first_x3n.png} 
%  \end{minipage}
%  \begin{minipage}{0.32\textwidth}
%    \includegraphics[scale=0.23]{first_x3n.png} 
%  \end{minipage} 
%\caption{TEST PLOT: Estimated Shapley curves for vehicle weight in dependence of weight (in lbs) for a vehicle length of $190$ inches and $190$ horsepower. %The time periods are 2001 - 2007 (top left), 2008 - 2013 (top right), 2014 - 2020 (bottom left) and pooled (bottom right).}
%\label{lbs_slice}
%\end{figure}

   % \item 3x3 slice plots \\
    Next, we illustrate the estimated confidence intervals for the Shapley Curves over the support of variable $j$. Therefore we fix the remaining variables at the median. The results of this exercise are shown in Figure \ref{lbs_slice} for vehicle weight, in Figure \ref{hp_slice_shap} for horsepower, and in Figure S4 in the supplementary material for vehicle length. This type of analysis is not to be mistaken with the interpretation of confidence bands. For each of these \textit{`slice plots'} we illustrate $\widehat{\phi}_j(x)$ against $X_j$. 
    Figure \ref{hp_slice_shap} leads to the following economics insight. Across the time domain, we infer that horsepower, in general, is contributing less to an increased price as we move from the first to the second and third time interval. This effect is especially prominent as we compare the first and second time interval. The reason is that as technology develops over time, it is less costly for car manufacturers to produce vehicles with higher horsepower.

    %xgboost + KernelSHAP + variable independence + no subsampling
   Further, we include estimated KernelSHAP values, as well as a smoothed curve of these values in Figure 1. KernelSHAP assumes independence between the variables and does not include subsampling of subsets in the sense of a computational approximation. As an estimator, we use \textit{XGBoost} with default hyperparameter settings. To make the comparison between Shapley curves and smoothed KernelSHAP values as fair as possible, we estimate KernelSHAP for Horsepower, only on observations with realizations of Weight and a Length lying in an interval around the median. It shows that the Shapley curve for horsepower and the smoothed SHAP values result in different curves in such a scenario. This is not surprising, as the assumption of variable independence can be practically misleading. Further, the smoothed SHAP curves are not accompanied by confidence intervals.

    %Further, we include estimated SHAP values, as well as a smoothed curve of these values. In order to make the comparison between Shapley curves and smoothed SHAP values as fair as possible, we estimate SHAP values for Horsepower, only on observations with realizations of Weight and a Length lying in an interval around the median. It shows that the Shapley curve for horsepower and the smoothed SHAP values result in similar curves in such a scenario. However, the smoothed SHAP curve is not accompanied by confidence intervals.

    The first row of Figure \ref{lbs_slice} includes the Shapley curve for vehicle weight. It shows that a lighter weight contributes to a price decrease, accompanied by stagnation at around 4000 lbs and an increase shortly after. Further, the Shapley curve is stretched apart on the tails across the time intervals. 
    The estimated confidence intervals have a larger spread as we move to the tails of the variables. This is intuitive as we have more variation in price for the observations in this area. For example, a very heavy car with more than 5500 lbs can either be a Nissan NV Cargo Van or a Cadillac Escalade SUV. On the left tail, we observe insignificance for the third period. 

The second row in Figure \ref{lbs_slice} includes the KernelSHAP values estimated for each observation in the respective time interval. In addition, we fit a smoother on these values, to make a comparison to our Shapley curves. In contrast to Figure \ref{hp_slice_shap}, we are estimating the KernelSHAP for all observations.

%The second row in Figure \ref{lbs_slice} includes the SHAP values estimated for each observation in the respective time interval. In addition, we fit a smoother on these values, to make a comparison to our Shapley curves. In contrary to Figure \ref{hp_slice_shap}, we are estimating the SHAP values for all observations. This illustrates the original approach of SHAP.

Further, we observe that the variable vehicle length, Figure S3 in supplementary material, mostly does not significantly contribute to the price prediction after a length of approximately 230 inches for the first and second time period. However, we observe a downward trend in the prices for longer cars the more recent the time interval is.

\section{Conclusion}
\label{section:conclusion}

This paper analyzes Shapley curves as a local measure of variable importance in a nonparametric framework. We give a rigorous definition of Shapley curves on the population level. As for estimation, we discuss two estimation strategies, the component-based approach, and the integration-based approach. We prove that both estimators globally converge with the nonparametric rate of \citet{stone1982optimal} to the population counterpart. Asymptotic normality is obtained for both estimators. We show that the asymptotic variance is identical for both approaches. However, the integration-based approach is accompanied by an inflated asymptotic bias. In our simulation exercise, this difference is visible in the results for the MISE. The advantage of the integration-based estimator is that only the pilot estimator is required to estimate the components. 
We proposed a tailored wild bootstrap procedure, which we proved to be consistent. Empirically, it results in good finite sample coverage. Under the assumption of an additive model a one-dimensional convergence rate results for Shapley curves by \citet{stone1985additive}.
For an extensive vehicle price data set we show that Shapley curves are a useful tool for the practitioner to gain insight into pricing and the importance of vehicle characteristics. The estimated confidence intervals enable us to distinguish between significant and non-significant sections of the variables. Building on our results, empirical researchers as well as practitioners can conduct statistical inference for Shapley curves.

\section*{Acknowledgements}
This research is supported by the German Academic Scholarship Foundation; Deutsche Forschungsgemeinschaft via SFB 1294 and IRTG 1792, Humboldt-Universität zu Berlin; the Yushan Fellowship; the Czech Science Foundation's grant no. 19-28231X / CAS: XDA 23020303 and the IDA Institute for Digital Assets, ASE, Bucharest.

% Acknowledgements should go at the end, before appendices and references
% \section{Acknowledgements}
% This research is supported by the Deutsche Forschungsgemeinschaft via IRTG 1792 “High Dimensional Nonstationary Time Series”, Humboldt-Universität zu Berlin; the European Union’s Horizon 2020 research and innovation program “FIN-TECH: A Financial supervision and Technology compliance training programme”under the grant agreement no. 825215 (Topic: ICT-35-2018, Type of action: CSA); the European Cooperation in Science \& Technology COST Action grant CA19130 - Fintech and Artificial Intelligence in Finance - Towards a transparent financial industry; the Yushan Scholar Program of Taiwan; and the Czech Science Foundation’s grant no. 19-28231X / CAS: XDA 23020303.

%\clearpage
\appendix

\clearpage
\centerline{\textbf{SUPPLEMENTAL: APPENDICES}}
  \medskip
} 
\bigskip
\bigskip

\spacingset{1.5} % DON'T change the spacing!

This supplementary document is organized as follows. In Section \ref{proofs} we provide proofs of Proposition 1, Theorem 2, Proposition 3, and Theorem 5. Intuition on the proposed bootstrap procedure for finite samples is given. In Section \ref{simulation} we give details on the simulation procedure and include supporting tables and figures of the simulation exercise as well as the empirical application.

%\clearpage
\appendix
\setcounter{equation}{0}
\renewcommand\theequation{S\arabic{equation}}

\section{Proofs}\label{proofs}
\subsection{Proof of Proposition 1}
\label{glob_proof}
\begin{proof}
Using the weighted sum representation (8) we write
\begin{align*}
\widehat{\phi}_j(x)-\phi_j(x) = \sum_{s \subseteq N} \omega_{j,s} \left\{ \widehat{m}_s(x_s) - m_s(x_s)  \right\}.
\end{align*}
%jensens
Taking the squared error and applying Jensen's inequality gives
\begin{align*}
\left\{\widehat{\phi}_j(x) - \phi_j(x)\right\}^2 &= \left[\sum_{s \subseteq N} \omega_{j,s} \left\{ \widehat{m}_s(x_s) - m_s(x_s)  \right\}\right]^2 \\
&\leq \sum_{s \subseteq N} \omega_{j,s}^2 \left\{ \widehat{m}_s(x_s) - m_s(x_s)\right\}^2.
\end{align*}
Next, we integrate over $\mathcal{X}$ resulting in 
\begin{align*}
\int_{\mathcal{X}} \left\{\widehat{\phi}_j(x)-\phi_j(x)\right\}^2 dx \leq \sum_{s \subseteq N} \omega_{j,s}^2 \int_{\mathcal{X}} \left\{ \widehat{m}_s(x_s) - m_s(x_s)\right\}^2 dx.
\end{align*}
Finally, taking the expectation gives the MISE
\begin{align*}
{\sf E}\left[ \int_{\mathcal{X}} \left\{\widehat{\phi}_j(x)-\phi_j(x)\right\}^2 dx \right] &\leq \sum_{s \subseteq N} \omega_{j,s}^2 {\sf E}\left[ \int_{\mathcal{X}} \left\{ \widehat{m}_s(x_s) - m_s(x_s)\right\}^2 dx \right] \\
&= \sum_{s \subseteq N} \omega_{j,s}^2 \text{MISE}\left\{ \widehat{m}_s(x_s),  m_s(x_s)\right\}.
\end{align*}
%stone
By Assumptions 1, 2 and 3, $m_s(x_s)$ belongs to the class of $|s|$-dimensional twice continuously differentiable functions.
By choosing the bandwidth as $h_s\sim n^{-\frac{1}{4+|s|}}$ and invoking Theorem 2 of \citet{fan1993local} one sees that the estimator achieves the minimax rate of \citet{stone1982optimal}:
\begin{align}
    \text{MISE} \left\{ \widehat{m}_s(x_s),  m_s(x_s)\right\} = \mathcal{O}\left(n^{-\frac{4}{4+|s|}}\right). \label{crate}
\end{align}
Therefore,
\begin{align*}
{\sf E}\left[ \int_{x} \left\{\widehat{\phi}_j(x)-\phi_j(x)\right\}^2 dx \right] &\leq \sum_{s \subseteq N} \omega_{j,s}^2 \mathcal{O}\left(n^{-\frac{2p}{4+|s|}}\right)\\
&= \mathcal{O}\left(n^{-\frac{4}{4+d}}\right) + {\scriptstyle\mathcal{O}}\left(n^{-\frac{4}{4+d}}\right).
\end{align*}
\end{proof}

\subsection{Proof of Theorem 2}
\label{proof_comp_asy}
\begin{proof}
Recall that $h_m\sim n^{-\frac{1}{4+d}}$ is the optimal bandwidth of the $d$-dimensional model and let $\frac{1}{d}$ denote the corresponding weight. From the proof of Proposition 1 it follows that
\begin{align*}
\widehat{\phi}_j(x) - \phi_j(x) = \frac{1}{d} \left\{\widehat{m}(x) - m(x)\right\} + {\scriptstyle\mathcal{O}}_p\left(n^{-\frac{2}{4+d}}\right).
\end{align*}
Asymptotic normality follows by Assumptions 1, 2, and 3 and the Lindeberg-Feller Central Limit Theorem (CLT), 
\begin{align*}
\sqrt{nh_m^d}\left\{\widehat{\phi}_j(x) - \phi_j(x)\right\} \stackrel{\mathcal{L}}{\rightarrow}N\left(B(x),V(x)\right),
\end{align*}
where the asymptotic bias $B(x)$ and asymptotic variance $V(x)$ are given in Theorem 2.1 of \citet{ruppert1994multivariate},
\begin{align*}
 B(x)&=\frac{1}{d} \frac{\mu_2(k)}{2} \sum^d_{j=1} \frac{\partial^2 m(x)}{\partial^2 x_j}, \\
 V(x)&=\frac{1}{d^2} ||k||^2_2 \frac{\sigma^2(x)}{f(x)}.
\end{align*}

\end{proof}

\subsection{Proof of Proposition 3}

\begin{proof}
To prove the consistency of the bootstrap procedure, we need to show that
\begin{align}
       \bigg \lvert P^{Y \mid X}\left[\sqrt{n h_m^d}\left\{\widehat{\phi}_j(x)-\phi_j(x)\right\}<z\right]-  \Phi_{B,V}(z) \bigg| {\rightarrow} 0, \label{bs1}
\end{align}
and
\begin{align}
       \bigg \lvert P^{*}\left[\sqrt{n h_m^d}\left\{\widehat{\phi}^{*}_j(x)-\widehat{\phi}_{g,j}(x)\right\}<z\right]-  \Phi_{B,V}(z) \bigg| {\rightarrow} 0, \label{bs2}
\end{align}
where $\Phi_{B,V}(z)$ denotes a normal distribution with mean $B(x)$ and covariance $V(x)$ as defined in Theorem 2.
To show (\ref{bs1}), we follow the proof of Lemma 1 of \citet{hardle1991bootstrap} by noting that,
    \begin{align}
     \sqrt{nh^d_m}\left\{\widehat{\phi}_j(x)-\phi_j(x)\right\} &= \sqrt{nh^d_m} \frac{1}{d} \left\{ \widehat{m}(x) - m(x) \right\} + {\scriptstyle\mathcal{O}}_p\left( 1 \right) \label{bsana}  \\
&= L_n + {\scriptstyle\mathcal{O}}_p(L_n) \nonumber,
    \end{align}
  where $L_n=\frac{1}{d} \sqrt{nh^d_m} \left[\frac{1}{n} \sum^n_{i=1} \frac{K_h(x-X_i)\left\{ Y_i - m(x) \right\}}{f(x)}\right]$.
Denote $W^{'}_i(x) = \sqrt{\frac{h^d_m}{n}} \frac{K_h(x-X_i)}{f(x)}$, then we can further decompose $L_n$ into a bias and a variance term, $L_n=B_n+V_n$ with
\begin{align*}
    B_n&=\sum_{i=1}^n W_i^{'}(x)\left\{m(X_i)-m(x)\right\},\\
    V_n&=\sum_{i=1}^n W_i^{'}(x)\varepsilon_i.
\end{align*}
While we have $B_n\stackrel{p}{\rightarrow}B(x)$, looking at the second and third moment of $V_n$, conditional on $X_1,\ldots,X_n$,
\begin{align*}
    S_{2n} = \sum^{n}_{i=1}\text{Var}\left\{W^{'}_i(x) \varepsilon_i \lvert X_1,\ldots,X_n \right\}= \mathcal{O}(1)
\end{align*}
and by Assumption 4,
\begin{align*}
    S_{3n} = \sum^{n}_{i=1}{\sf E}\left\{ \lvert W^{'}_i(x) \varepsilon_i \lvert^3  \lvert X_1,\ldots,X_n \right\}=\mathcal{O}\left(\frac{1}{\sqrt{nh}}\right).
\end{align*}
We can apply Theorem 3 on page 111 in \citet{petrov2022sums}, the Esseen's inequality for arbitrary independent random variables, to bound (\ref{bs1}) by noting, 
% By following \citet{hardle1991bootstrap} we know that
% \begin{align}
% S_{2n} = \mathcal{O}(h^{-1}) \label{s2n}
% \end{align}
% and 
% \begin{align}
% S_{3n} = \mathcal{O}\left(\frac{1}{\sqrt{nh}}\right). \label{s3n}
% \end{align}
% From (\ref{s2n}) and (\ref{s3n}) it follows that
\begin{align*}
\frac{S_{3n}}{S^{\frac{3}{2}}_{2n}} = {\scriptstyle\mathcal{O}}(1)\ \text{a.s.}
\end{align*}
% such that equation (\ref{bs1}) follows. In a similar way equation (\ref{bs2}) can be proven. 
To prove (\ref{bsana}), we have to obtain a similar bound for the bootstrap version of the estimator. Note that
\begin{align*}
     \sqrt{nh^d_m}\left\{     \widehat{\phi}^{*}_j(x)-\widehat{\phi}_{g,j}(x)         \right\} &= \sqrt{nh^d_m} \frac{1}{d} \left\{ \widehat{m}^{*}(x) - \widehat{m}_g(x) \right\} + {\scriptstyle\mathcal{O}}_p\left( 1 \right), 
\end{align*}
which is a similar starting argument as in the proof of (\ref{bs2}). The remaining steps follow by Lemma 2 of \citet{hardle1991bootstrap}. 

\end{proof}

\subsection{Intuition on the wild bootstrap adjustment}
It is quite useful to provide evidence for the performance of the bootstrap algorithm. This will provide some intuition for including the lower-order terms. Recall that $\widehat{\phi}^{*}_j(x) =  \sum_{s \subseteq N} \omega_{j,s} \widehat{m}^{*}_{s}(x_{s})$ and $\widehat{\phi}_{j,g}(x) =  \sum_{s \subseteq N} \omega_{j,s} \widehat{m}_{s,g}(x_{s})$. Let $W_{s,i}(x_s)$ denote the local linear weights for $\widehat{m}_s(x_s)$. By neglecting the bias term, it can be shown that
%Let the {\color{blue}local linear weight} for $\widehat{m}_s(x_s)$ be $W_{s,i}(x_s)=\frac{K_h(x_s-X_{s,i})}{f_s(x_s)}$. {\color{blue}By simple algebra}, it can be shown that
\begin{align}
    \widehat{\phi}_j(x) - \phi_j(x) &\approx \sum^n_{i=1} \left\{\sum_{s \subseteq N} W_{s,i}(x_s) \omega_{j,s} \varepsilon_{s,i}\right\} \label{term1} \\
    \widehat{\phi}^{*}_j(x) - \widehat{\phi}_{j,g}(x)    &\approx \sum^n_{i=1} \left\{\sum_{s \subseteq N} W_{s,i}(x_s) \omega_{j,s} \widehat{\varepsilon}_{s,i}\right\} V_i. \label{term2}
\end{align}
The conditional variances of (\ref{term1}) and (\ref{term2}) are given by
\begin{align}
    \text{Var}\left\{\widehat{\phi}_j(x) - \phi_j(x) \lvert X\right\} &= \sum^{n}_{i=1} \sum_{s\in\mathcal{S}} \sum_{s^{'}\in\mathcal{S}} \omega_{j,s} \omega_{j,s^{'}} W_{s,i}(x_s) W_{s^{'},i}(x_s) {\sf E}(\varepsilon_{s,i} \varepsilon_{s^{'},i} \lvert X) \label{car1} \\
  \text{Var}\left\{\widehat{\phi}^{*}_j(x) - \widehat{\phi}_{j,g}(x) \lvert X\right\} &= \sum^{n}_{i=1} \sum_{s\in\mathcal{S}} \sum_{s^{'}\in\mathcal{S}} \omega_{j,s} \omega_{j,s^{'}} W_{s,i}(x_s) W_{s^{'},i}(x_s) {\sf E}(\widehat{\varepsilon}_{s,i} \widehat{\varepsilon}_{s^{'},i}\lvert X) {\sf E}(V^2_i). \label{car2}
\end{align}
Asymptotically, the leading terms in the conditional variance expressions are
\begin{align*}
    \text{Var}\left\{\widehat{\phi}_j(x) - \phi_j(x) \lvert X\right\} &= \sum^{n}_{i=1} \frac{1}{d^2} W_{m,i}^2(x) {\sf E}(\varepsilon^2_i \lvert X) + {\scriptstyle\mathcal{O}}\left(n^{-\frac{4}{4+d}}\right)\\
  \text{Var}\left\{\widehat{\phi}^{*}_j(x) - \widehat{\phi}_{j,g}(x) \lvert X\right\} &= \sum^{n}_{i=1} \frac{1}{d^2} W_{m,i}^2(x) {\sf E}(\widehat{\varepsilon}^2_{i}\lvert X) + {\scriptstyle\mathcal{O}}\left(n^{-\frac{4}{4+d}}\right).
\end{align*}
While not relevant for the asymptotic results, the smaller order terms in (\ref{car1}) and (\ref{car2}) can play an important role in finite samples. Including these terms will lead to a more accurate reflection of the variance by the bootstrap procedure. As we see, the variance terms (\ref{car1}) and (\ref{car2}) are similar, given that ${\sf E}(\varepsilon_{s,i} \varepsilon_{s^{'},i} \lvert X) \approx {\sf E}(\widehat{\varepsilon}_{s,i} \widehat{\varepsilon}_{s^{'},i}\lvert X)$ and ${\sf E}(V^2_i)=1$.

%For the bias term it holds that
%\begin{align*}
%    {\sf E} \left\{\widehat{\phi}_j(x) - \phi_j(x)\right\} = {\sf E}  \left\{\widehat{\phi}^{*}_{j}(x) - \widehat{\phi}_{j,g}(x)\right\},
%\end{align*}
%since we know that asymptotically 
%\begin{align*}
%    {\sf E}\left\{\widehat{m}_{s}(x_s) - m_s(x_s)\right\} = {\sf E}\left\{\widehat{m}^{*}_s(x_s) - \widehat{m}_{s,g}(x_s)\right\},\text{ for all %} s\in\mathcal{S}.
%\end{align*}

\subsection{Proof of Theorem 5}
\label{proof_integ_asy}
\begin{proof}
Let $h_m \sim n^{-\frac{1}{4+d}}$ and recall Assumptions 1, 2 and 3 to obtain
\begin{align*}
    \text{MISE} \left\{ \widetilde{m}(x),  m(x)\right\} = \mathcal{O}\left(n^{-\frac{4}{4+d}}\right). 
\end{align*}
The global convergence of $\widetilde{m}_s(x_s)$ results directly from the continuous mapping theorem (CMT), such that
\begin{align}
    \text{MISE} \left\{ \widetilde{\phi}_j(x),  \phi_j(x)\right\} = \mathcal{O}\left(n^{-\frac{4}{4+d}}\right). 
\end{align}

Next, we prove the asymptotic normality of $\widetilde{m}_s(x_s)$. Similar to the proof of Theorem 2, the weighted sum representation leads to
\begin{align*}
\widetilde{\phi}_j(x) - \phi_j(x) &= \sum_{s \subseteq N} \omega_{j,s}\left\{\widetilde{m}_s(x_s) -  m_s(x_s)\right\}. 
\end{align*}
By Assumptions 1,2 and 3, the Lindeberg-Feller central limit theorem and the continuous mapping theorem it follows that
\begin{align*}
 \sqrt{nh_m^d}\sum_{s \subseteq N} \omega_{j,s}\left\{\widetilde{m}_s(x_s) -  m_s(x_s)\right\}
 \stackrel{\mathcal{L}}{\rightarrow}N\left(B_{int}(x),V(x)\right),
\end{align*}
where the asymptotic bias $B_{int}(x)$ and the asymptotic variance $V(x)$ are
\begin{align*}
 B_{int}(x)&=\frac{\mu_2(k)}{2} \sum_{s \subseteq N} \omega_{j,s} \left\{ \sum^d_{j=1} \int_{X_{-s}} \frac{\partial^2 m(x)}{\partial^2 x_j} f_{X_{-s}|X_s}(x_{-s}|x_s)dx_{-s}\right\}, \\
 V(x)&=\frac{1}{d^2} ||k||^2_2 \frac{\sigma^2(x)}{f(x)}.
\end{align*}
The bias accumulates as the conditional mean for each subset converges at the same rate to the population counterpart, whereas the variance is the same as for the component-based estimator. We prove both in the following.

First, we derive the bias of the integration-based estimator.
By using the weighted sum representation we get
\begin{align}
\label{bias:shap}
{\sf E}\left\{ \widetilde{\phi}_j(x) \right\} - \phi_j(x) = \sum_{s \subseteq N} \omega_{j,s} \left[ {\sf E}\left\{  \widetilde{m}_s(x_s) \right\}-m_s(x_s) \right].
\end{align}
By Assumptions 1,2 and 3 the bias of the integration-based estimator of the subset $s$, $\widetilde{m}_s(x_s)$, is given as
\begin{align}
\label{bias:comp}
{\sf E}\left\{  \widetilde{m}_s(x_s) \right\}-m_s(x_s) \approx  h_m^2 \mu_2(k) \frac{1}{2} \left\{ \sum^d_{j=1} \int_{X_{-s}} \frac{\partial^2 m(x)}{\partial^2 x_j} d F_{X_{-s}|X_s}(x_{-s}|x_s) \right\}. 
\end{align}
The result follows as we insert (\ref{bias:comp}) into (\ref{bias:shap}).

Next, we derive the asymptotic variance of the integration-based estimator. Under Assumptions 1,2 and 3 we show that:
\begin{align}
    \text{Var}\left\{\widetilde{\phi}_j(x)\right\} = \frac{1}{d^2} \text{Var}\left\{\widetilde{m}(x)\right\} + {\scriptstyle\mathcal{O}}\left(\frac{1}{nh_m^d}\right) \label{proof_var}. 
\end{align}
Invoking \citet{linton1995kernel} one derives the asymptotic variance of the integration-based estimator for a given component function. The key idea is to consider the variance of the internal estimator (estimator with known density) and show that the difference between this internal estimator and the local linear estimator is of smaller order.
Note that the proof also works for dimension $d>2$. The reason is that the difference between the internal estimator and the local linear estimator is of small order for any dimension $d$, namely $\mathcal{O}_p(n^{-1}h_m^{-d})+{\scriptstyle\mathcal{O}}_p(h_m^d)$.

As a consequence the variance of $\widetilde{m}_s(x_s)$, for $|s|<d$, is given by
\begin{align*}
    \text{Var}\left\{\widetilde{m}_s(x_s)\right\} &\approx \frac{1}{nh_m^{|s|}} \|k\|^2_2\int \frac{\sigma^2(x)f^2_{X_{-s}|X_s}(x_{-s}|x_s)}{f(x)}dx_{-s}\\
    &= \mathcal{O}\left(\frac{1}{nh_m^{\lvert s \rvert}}\right) = {\scriptstyle\mathcal{O}}\left(\frac{1}{nh_m^d}\right).
\end{align*}
Equation (\ref{proof_var}) follows directly since the variance of the lower dimensional components as well as the covariance terms converge at a faster rate than the variance of the $d$-dimensional model.

\end{proof}

\section{Simulation Details, Tables and Plots}\label{simulation}

%BS Setup
The coverage probability in Table 2 is estimated for $\alpha=0.05,0.1$ and $0.15$ for increasing sample size.
First, we simulate data following (15) with additional Gaussian noise. The proposed bootstrap procedure is conducted as described in Section 3.1. We estimate $\widehat{\phi}^b_j(x_s)$ at the point $(-0.5,-0.5)$. 
The procedure is repeated in $B=1000$ replications, where we estimate the component-based Shapley curve for each replication $b=1,\ldots,B$. We choose the bandwidth $g_s=h_s\times \log (\log (n) ) \times 4$ such that $\frac{h_s}{g_s} \rightarrow 0$ as $n \rightarrow \infty$ for all subsets $s\in\mathcal{S}$.

%On MISE calcualtion
The integral within the MISE is approximated with the \textit{cubature} R package, which adaptively calculates the multidimensional integral over hypercubes.

\begin{table}[H]
\spacingset{1.2}
\caption{MISE of Shapley Curves for component-based estimator $\widehat{\phi}_j$ and integration-based estimator $\widetilde{\phi}_j$ for each variable. An additive and non-linear DGP is used for $\rho=0$ and $\rho=0.8$. The error terms follow $\varepsilon \sim t(5)$. }\label{tab2}
\centering
%\hspace*{-1.1cm}
\setlength{\tabcolsep}{8pt}
\begin{tabular}{ r r r r r r r r r }
 DGP & $\rho$ & $n$ & $\widehat{\phi}_1$ & $\widetilde{\phi}_1$ & $\widehat{\phi}_2$ & $\widetilde{\phi}_2$ &$\widehat{\phi}_3$ & $\widetilde{\phi}_3$  \\
\hline
\multirow{8}{*}{Additive} & \multirow{4}{*}{$0$} & 300 & \bftab{10.99} & 20.05 & 10.17 & \bftab{9.71} & 3.06 & \bftab{0.44} \\ 
& &  500 & \bftab{5.96} & 7.95 & \bftab{6.38} & 8.40 & 2.02 & \bftab{0.28} \\ 
& & 1000 & \bftab{3.71} & 4.41 & \bftab{4.30} & 5.41 & 1.09 & \bftab{0.16} \\ 
& & 2000 & \bftab{2.26} & 2.65 & \bftab{2.63} & 3.34 & 0.62 & \bftab{0.09} \\  
\cline{2-9} 
& \multirow{4}{*}{$0.8$} & 300 & \bftab{9.13} & 10.57 & \bftab{9.88} & 11.45 & 3.10 & \bftab{1.81} \\ 
 & & 500 & \bftab{6.39} & 7.14 & \bftab{6.76} & 7.95 & 2.12 & \bftab{1.19} \\ 
 & & 1000 & \bftab{4.32} & 4.56 & \bftab{4.38} & 5.03 & 1.44 & \bftab{0.73} \\ 
 & & 2000 & \bftab{2.84} & 2.91 & \bftab{2.86} & 3.26 & 0.93 & \bftab{0.45} \\ 
\hline
\multirow{8}{*}{Non-linear} & \multirow{4}{*}{$0$} & 300 & \bftab{11.48} & 17.13 & \bftab{15.25} & 20.06 & 2.37 & \bftab{0.57} \\ 
 & & 500 & \bftab{7.25} & 10.04 & \bftab{8.43} & 10.29 & 1.03 & \bftab{0.32} \\ 
 & & 1000 & \bftab{4.15} & 5.73 & \bftab{4.93} & 6.04 & 0.56 & \bftab{0.18} \\ 
 & & 2000 & \bftab{2.46} & 3.41 & \bftab{2.85} & 3.62 & 0.34 & \bftab{0.11} \\ 
\cline{2-9} 
& \multirow{4}{*}{$0.8$} & 300 & \bftab{12.60} & 14.84 & \bftab{14.82} & 17.38 & 3.61 & \bftab{2.44} \\ 
& & 500 & \bftab{9.35} & 10.35 & \bftab{9.77} & 12.24 & 2.64 & \bftab{1.71} \\ 
& & 1000 & 6.07 & \bftab{6.06} & \bftab{5.98} & 7.53 & 1.78 & \bftab{1.05} \\ 
& & 2000 & 4.04 & \bftab{3.82} & \bftab{3.77} & 4.73 & 1.22 & \bftab{0.68} \\ 
\hline
\end{tabular}
\end{table}

\begin{figure}[H] 
\begin{minipage}{0.32\textwidth}
    \includegraphics[scale=0.23]{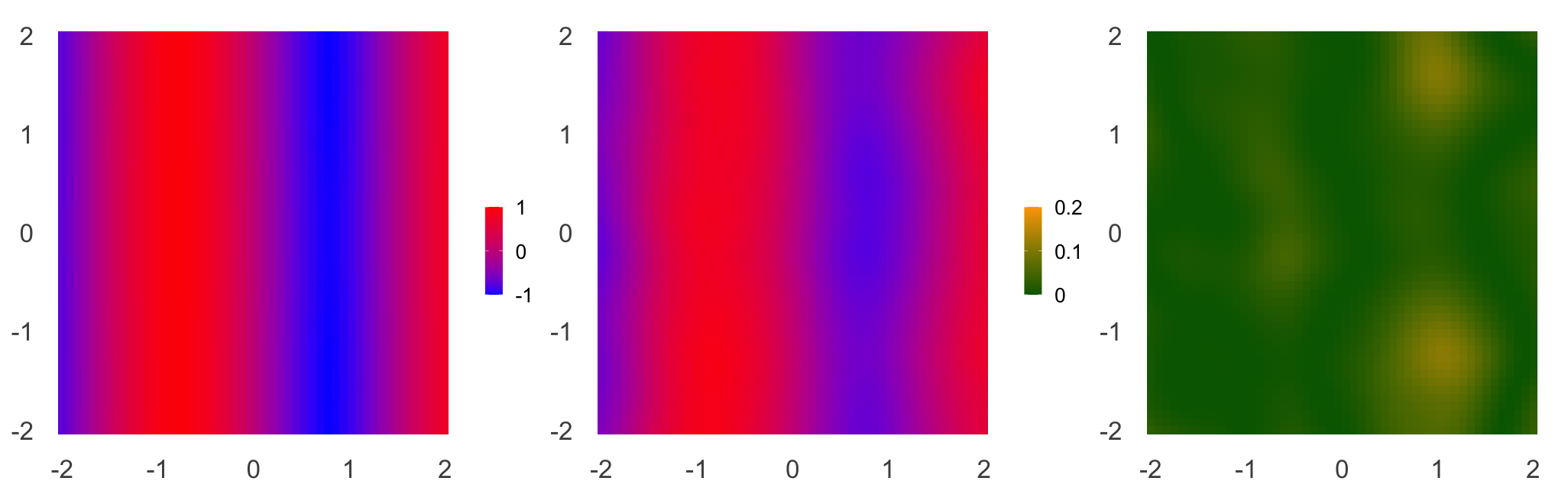} 
  \end{minipage} 
\caption{Heatmaps for $m(x) = -\text{sin}(2x_1) + \text{cos}(2x_2) + x_3 $ with Gaussian error terms with $n=2000$ for the first variable at $x_3=0$. Left: Population Shapley Curve. Centre: Componend-based estimated Shapley Curve. Right: Squared residuals between estimated and population curve.}
\label{heat_brown}
\end{figure}

\begin{table}[H]
\spacingset{1.2}
\caption{Summary statistics}\label{sum_stats}
\centering
\begin{tabular}{rrrrrrr}
  \hline
 & Mean & Min & $q_{0.25}$ & $q_{0.5}$  & $q_{0.75}$ & Max \\ 
  \hline
Horsepower & 258 & 70 & 181 & 260 & 310 & 808 \\ 
  Weight [lbs] & 4233 & 1808 & 3322 & 3970 & 5042 & 8039 \\ 
  Length [in] & 199 & 106 & 180 & 192 & 219 & 290 \\ 
  Price [1000 USD] & 36.35 & 8.9 & 24.1 & 31.23 & 41.19 & 191.5 \\
   \hline
\end{tabular}
\end{table}

\begin{figure}[H]
\begin{minipage}{0.5\textwidth}
		\includegraphics[scale=0.35]{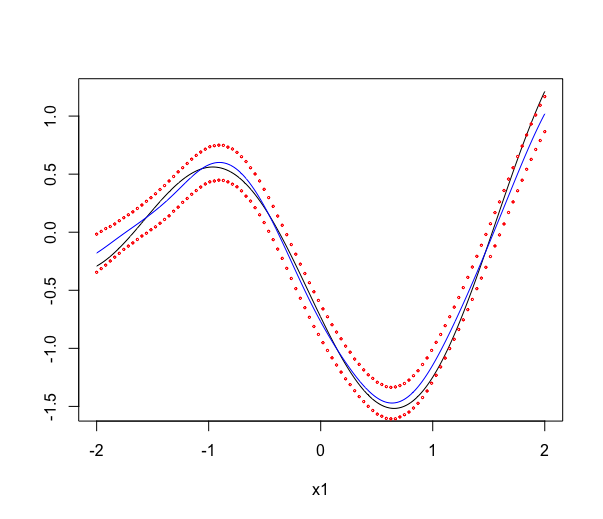}
\end{minipage}%
\begin{minipage}{0.5\textwidth}
\hspace{-0.5cm}
		\includegraphics[scale=0.35]{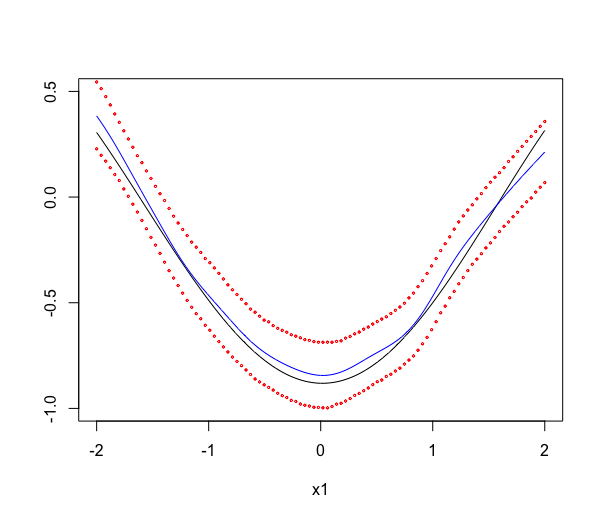}
\end{minipage}\\
\vspace{-1cm}
\caption{Population Shapley Curve (black) and component-based estimated Shapley Curve (blue) with $95\%$ bootstrap confidence intervals (red) for the first variable (left) and second variable (right). $x_1$ varies on [-2,2] and $x_2=-0.5$. The DGP 3 is used with Gaussian error terms. $B=10000$ bootstrap replications and $n=2000$.}
\label{slice_plot2}
\end{figure}

\begin{figure}[H] 
    \centering
    \includegraphics[scale=0.28]{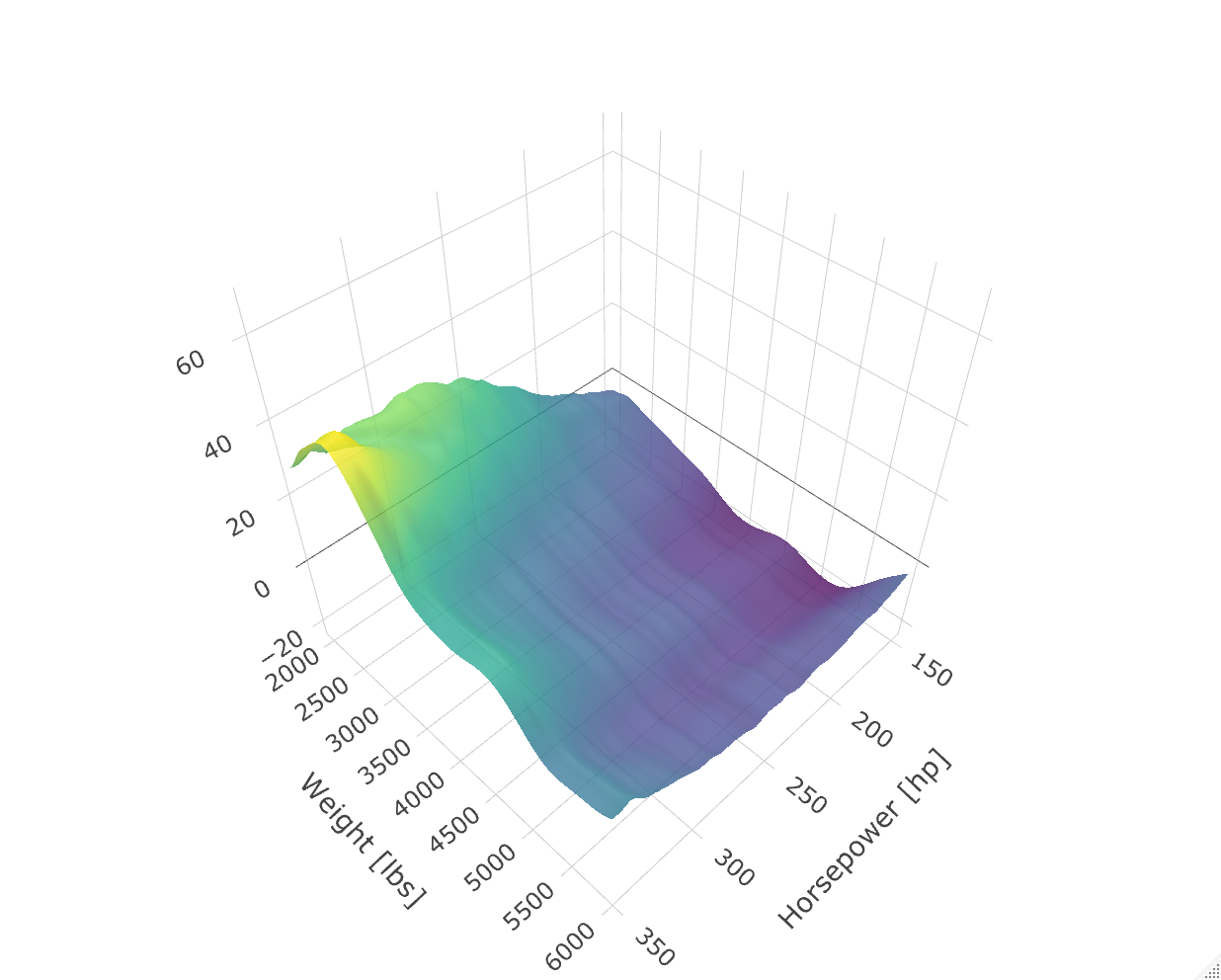}
    \caption{Estimated component-based Shapley curve for horsepower in dependence of vehicle weight (in lbs) and horsepower (in hp) for a vehicle length of $190$ inches for the pooled data set from the year 2001--2020.}
    \label{fig:my_label}
\end{figure}

%\begin{figure}[ht]
%  \begin{minipage}{0.5\linewidth}
%    \includegraphics[scale=0.18]{first_x1_3d.png} 
%    \vspace{-0.5cm}
%  \end{minipage}%%
%  \begin{minipage}{0.5\linewidth}
%    \hspace{-1.4cm}
%    \includegraphics[scale=0.18]{second_x1_3d.png} 
%  \end{minipage}
%  \begin{minipage}{0.5\linewidth}
%    \includegraphics[scale=0.18]{third_x1_3d.png} 
%  \end{minipage}%% 
%  \begin{minipage}{0.5\linewidth}
%     \hspace{-1.4cm}
%    \includegraphics[scale=0.18]{all_x1_3d.png} 
%  \end{minipage} 
%\caption{Estimated Shapley curves for horsepower in dependence of vehicle weight (in lbs) and horsepower (in hp) for a vehicle length of $190$ inches. The time periods are 2001 - 2007 (top left), 2008 - 2013 (top right), 2014 - 2020 (bottom left) and pooled (bottom right).}
%\label{3d_emp}
%\end{figure}    

\begin{figure}[ht] 
\begin{minipage}{0.32\textwidth}
    \includegraphics[scale=0.243]{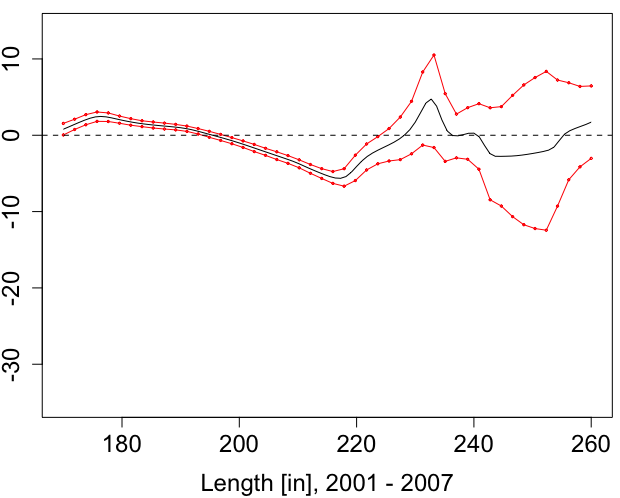} 
  \end{minipage}
  \begin{minipage}{0.32\textwidth}
    \includegraphics[scale=0.243]{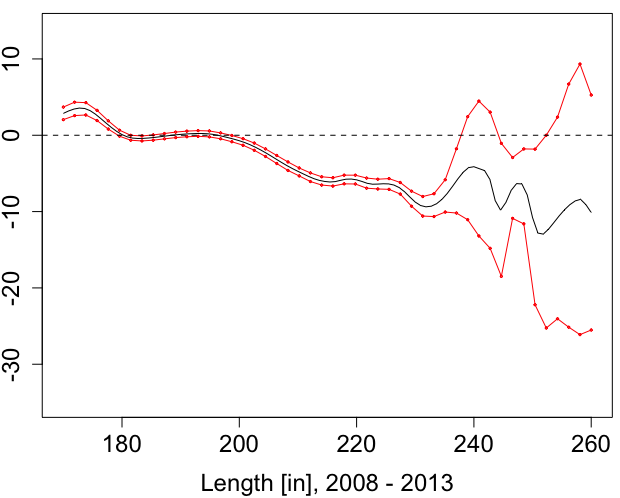} 
  \end{minipage}
  \begin{minipage}{0.32\textwidth}
    \includegraphics[scale=0.243]{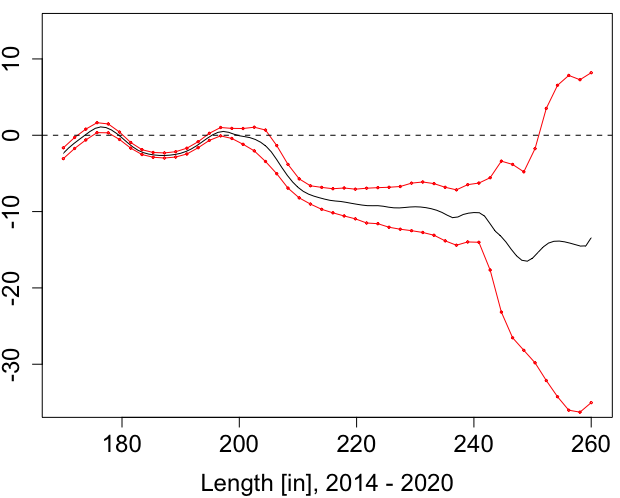} 
  \end{minipage} 
  \caption{Estimated component-based Shapley curves for vehicle length in dependence of length (in) for a vehicle weight of $3500$ pounds and horsepower of $190$. The time periods are 2001--2007, 2008--2013, and 2014--2020. }
\label{in_slice}
\end{figure}

\section{Further Simulations}
Motivated by \citet{hiabu2020random}, we consider two different simulation scenarios. Let the regressors be zero mean Gaussian with variance $\sigma^2=4$ and correlation $\rho$ of $0.8$ 
\begin{align*}
&\textbf{\text{DGP 4: Additive}}\ \ \ \ \ \ & m(x) &= \sum^d_{j=1} (-1)^{j} \sin{2 \pi x_j} \\
&\textbf{\text{DGP 5: Interactive}}\ \ \ 
& m(x) &= -2\sin(\pi x_1 x_2) - 2\sin(\pi x_1 x_3) + 2\sin(\pi x_2 x_3) \\
& & &\phantom{=} + \sum^d_{j=1} (-1)^{j} \sin{2 \pi x_j}. 
\end{align*}
For $M=3000$ Monte Carlo iterations, we calculate a discretized version of the MISE for the component-based estimator, namely the average MSE 
\begin{align*}
{\sf E}\left[ \frac{1}{n} \sum^n_{i=1} \left\{\widehat{\phi}_j(x_i)-\phi_j(x_i)\right\}^2 \right],
\end{align*}
where the average is calculated over the Monte Carlo iterations. The average MSE and the computational run time of a single Monte Carlo iteration are reported in Table \ref{tab10}.

\begin{table}[H]
\spacingset{1.2}
\caption{Average MSE and run time in seconds [s] and minutes [m] of Shapley Curves for the component-based estimator $\widehat{\phi}_j$ for the first variable and dimension $d=3,5,7,9$. The error terms follow $\varepsilon \sim N(0, 1)$. }\label{tab10}
\centering
\begin{tabular}{rrrrrrrrr}
  \hline
     & \multicolumn{4}{c}{Additive} & \multicolumn{4}{c}{Interactive}\\
  \hline
  $n$ & \multicolumn{8}{c}{MSE} \\\cmidrule{2-9}
 & 3 & 5 & 7 & 9 & 3 & 5 & 7 & 9 \\ 
  \hline
    300 & 0.44 & 0.83 & 1.09 & 1.27 & 0.65 & 1.02 & 1.19 & 1.29 \\
    500 & 0.36 & 0.74 & 1.01 & 1.18 & 0.53 & 0.95 & 1.15 & 1.23 \\ 
    1000 & 0.27 & 0.63 & 0.92 & 1.13 & 0.41 & 0.86 & 1.10 & 1.20 \\
    2000 & 0.20 & 0.54 & 0.84 & 1.06 & 0.31 & 0.77 & 1.06 & 1.15 \\ 
   \hline
  & \multicolumn{8}{c}{Time} \\\cmidrule{2-9}
   300 & 0.69s & 1.12s & 3.06s & 12.72s & 0.66s & 1.08s & 3.07s & 13.09s \\  
   500 & 0.84s & 1.81s & 6.47s & 29.87s & 0.91s & 1.90s & 6.25s & 29.87s \\
   1000 & 2.02s & 5.54s & 22.61s & 1.88m & 2.27s & 5.49s & 22.88s & 1.79m \\
   2000 & 7.34s & 19.30s & 1.51m & 7.55m & 7.70s & 21.81s & 1.49m & 7.21m \\
   \hline
\end{tabular}
\end{table}

\section{Further Empirical Results}
\begin{figure}[H] 
  \begin{minipage}{0.3\linewidth}
    \includegraphics[scale=0.23]{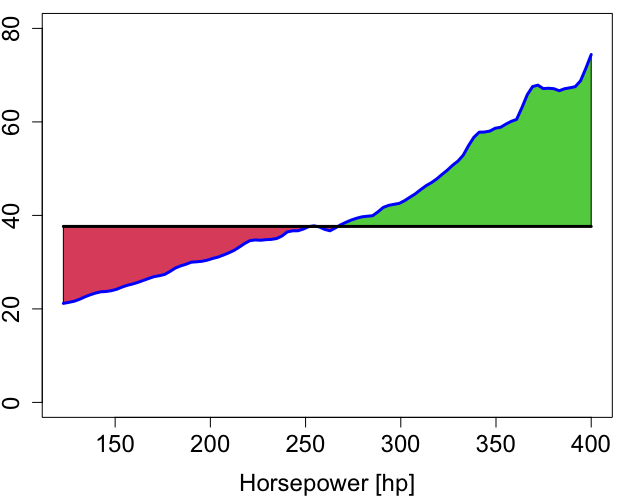} 
  \end{minipage}
\hfill
  \begin{minipage}{0.3\linewidth}
    \includegraphics[scale=0.23]{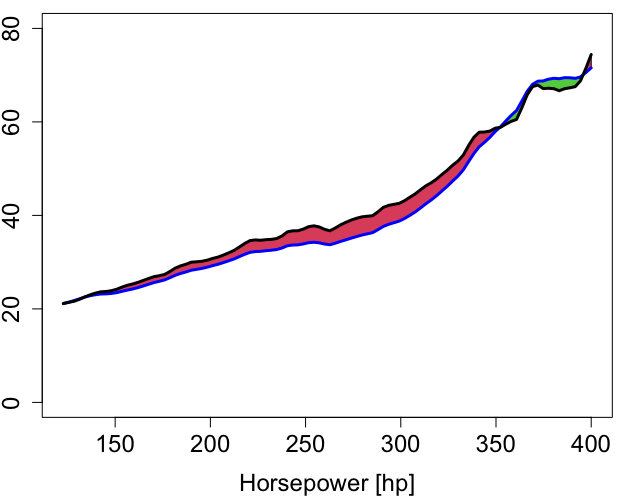} 
  \end{minipage}
  \hfill
  \begin{minipage}{0.3\linewidth}
    \includegraphics[scale=0.23]{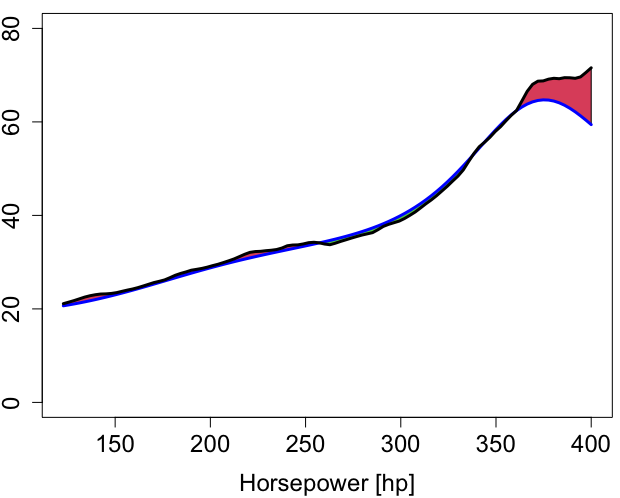} 
  \end{minipage}
  \begin{minipage}{0.3\linewidth}
    \includegraphics[scale=0.23]{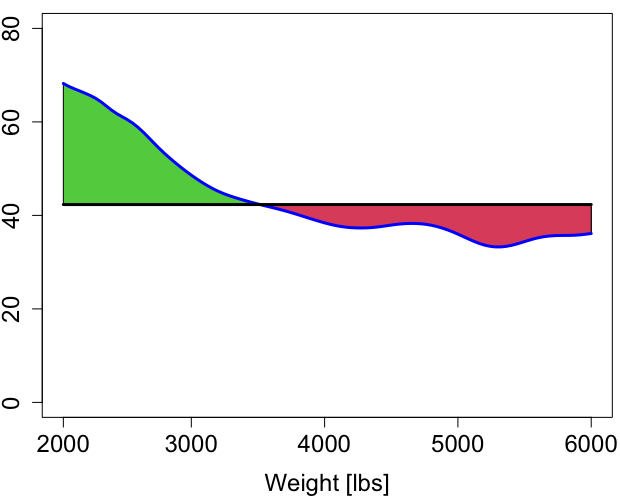} 
  \end{minipage} 
  \hfill
  \begin{minipage}{0.3\linewidth}
    \includegraphics[scale=0.23]{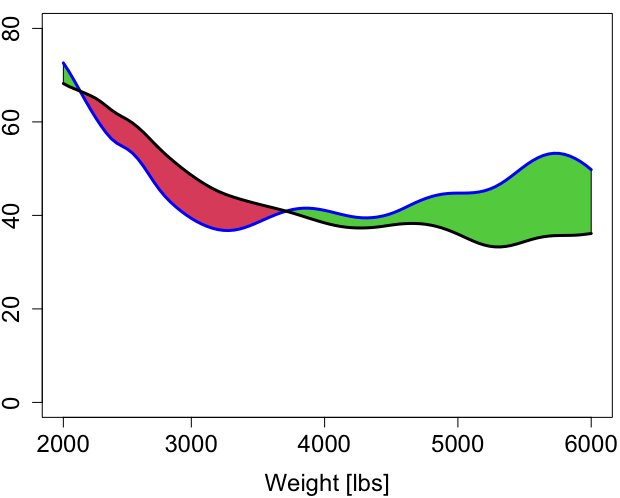} 
  \end{minipage}
  \hfill
  \begin{minipage}{0.3\linewidth}
    \includegraphics[scale=0.23]{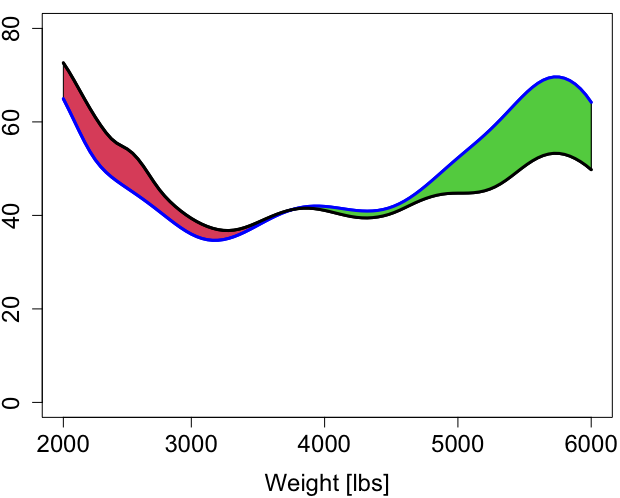} 
  \end{minipage}
    \begin{minipage}{0.3\linewidth}
    \includegraphics[scale=0.23]{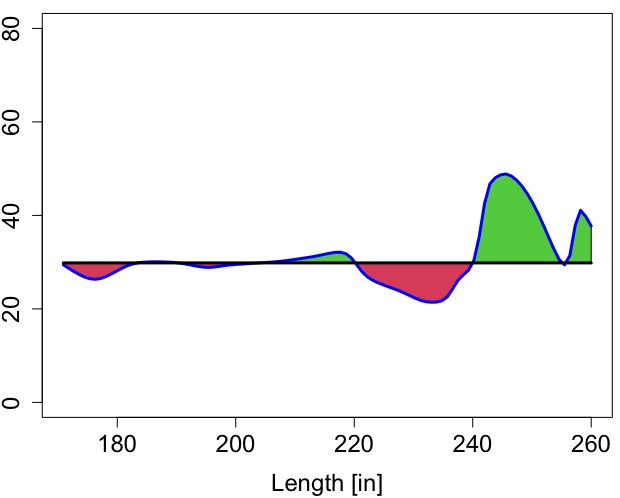} 
  \end{minipage}
  \hfill
    \begin{minipage}{0.3\linewidth}
    \includegraphics[scale=0.23]{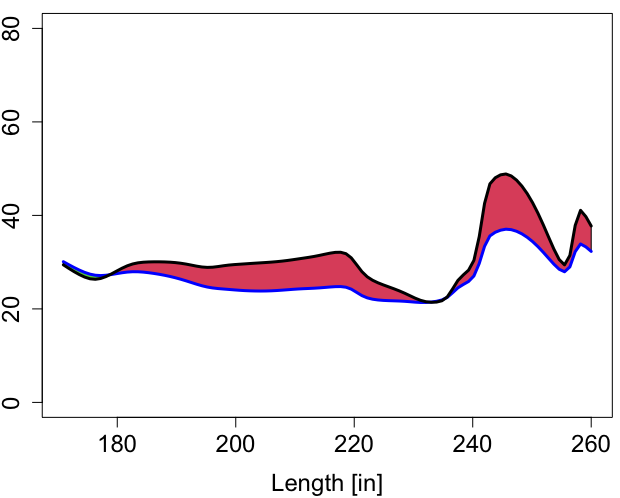} 
  \end{minipage}
  \hfill
    \begin{minipage}{0.3\linewidth}
    \includegraphics[scale=0.23]{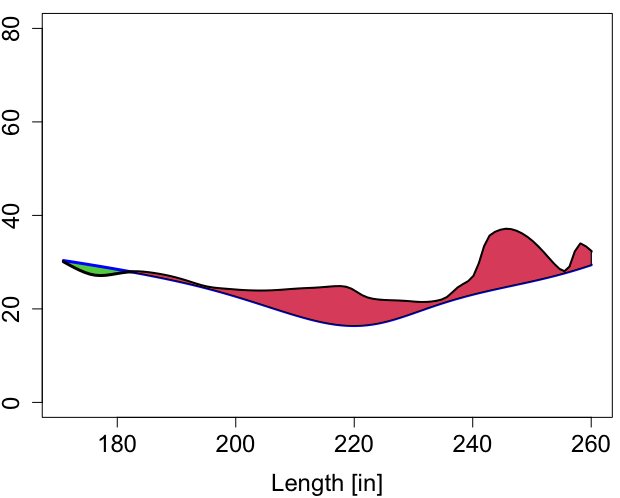} 
  \end{minipage}
  \caption{Cumulative Shapley Curves: The green area indicates an increase and the red area indicates a decrease in the price prediction of the updated cumulative Shapley curve (in blue).
Left panel shows $\bar{Y}$ in black and $\widehat{\phi}_1(x)+\bar{Y}$ in blue; Mid panel shows $\widehat{\phi}_1(x)+\bar{Y}$ in black and $\widehat{\phi}_1(x)+\widehat{\phi}_2(x) +\bar{Y}$ in blue; Right panel shows $\widehat{\phi}_1(x)+\widehat{\phi}_2(x) +\bar{Y}$ in black and $\widehat{\phi}_1(x) + \widehat{\phi}_2(x) + \widehat{\phi}_3(x) +\bar{Y}=\widehat{m}(x)$ in blue. The third plot in each row contains the conditional mean prediction in blue.
First row for vehicles with a weight of 3500 lbs and a length of 190 inches.
Second row for vehicles with a horsepower of 250 and a length of 190 inches.
Third row for vehicles with a horsepower of 250 lbs and weight of 3500 lbs.}
\label{seq_plot}
\end{figure}

%Two things can be observed:
We are interested in the cumulative price contribution of the variables at hand. Therefore we sequentially accumulate the Shapley curves to the unconditional mean prediction, such that the additional contribution of a variable to the prediction is visible. For instance, the green area in Figure \ref{seq_plot} indicates an increase in the price prediction, after adding up a certain variable. We call this the \textit{cumulative Shapley curves}, which are a function of a single variable of interest, with fixed remaining variables.

For example, the first row of Figure \ref{seq_plot} includes the cumulative Shapley curves as a function of horsepower for a car length of 190 inches and 3500 pounds. As we see, adding the curves for vehicle weight and length barely contributes to the prediction of price. In other words, their importance as a function of horsepower is negligible. 
The second row of Figure \ref{seq_plot} shows that as we include the second and third variables as a function of weight, we see that the price contribution mainly changes as we move closer to the tails. 
The first column of the said figure indicates how much more the addition of horsepower explains price differences in comparison to the average prediction. If we plot it as a function of horsepower, it contributes a lot, which means the area is relatively large.

A similar argument holds for the Shapley curve for horsepower, as a function of weight. However, for the third variable, the corresponding area is rather small. In general, we observe that even adding the Shapley curve for weight and length does not contribute a lot to the price prediction as a function of length.

A qualitative comparison can be made to the additive model. First, note that the first row suggests that the price contribution of weight and length does not depend on horsepower. Assume that the unknown data-generating process indeed follows an additive model as defined in Assumption 5. In that case Figure \ref{seq_plot} would have relatively large areas in the diagonal plots and an almost non-existent area in the off-diagonals. We conclude that this graphical analysis of sequential Shapley curves enables a heuristic distinction to the additive model.
%In case of an additive model with independent covariates,...

\clearpage

\bibliographystyle{apalike}

\bibliography{literature}

\end{document}